\documentclass[10pt,twocolumn,letterpaper]{article}
\usepackage[table]{xcolor}
\usepackage{3dv}

\usepackage{times}
\usepackage{amsfonts}
\usepackage{amsmath}
\usepackage{amssymb}
\usepackage{amsthm}
\usepackage{blindtext}
\usepackage{boldline}
\usepackage{booktabs}
\usepackage{caption}
\usepackage{colortbl}
\usepackage{comment}
\usepackage{cuted}
\usepackage{dsfont}
\usepackage{floatrow}
\usepackage{graphicx}
\usepackage{lipsum}
\usepackage{microtype}
\usepackage{multirow}
\usepackage{nicefrac}
\usepackage{pifont}
\usepackage{url}
\usepackage{wrapfig}
\usepackage[pagebackref=true,breaklinks=true,letterpaper=true,colorlinks,bookmarks=false]{hyperref}
\usepackage{cleveref}

\DeclareMathSymbol{\jj}{\mathalpha}{letters}{"11}
\newcommand{\bTheta}{\Theta}
\newcommand{\btheta}{\boldsymbol{\theta}}
\newcommand{\bbeta}{\boldsymbol{\beta}}
\newcommand{\bhj}{\mathbf{\hat{\jj}}}

\newcommand{\bphi}{\Phi}
\newcommand{\bI}{\mathbf{I}}
\newcommand{\todo}[1]{}

\newcommand{\cmark}{\ding{51}}%
\newcommand{\xmark}{\ding{55}}%

\definecolor{lightgray}{gray}{0.9}
\definecolor{SigmaColor}{rgb}{0.98,0.45,0.0}

\makeatletter
\DeclareRobustCommand\onedot{\futurelet\@let@token\@onedot}
\def\@onedot{\ifx\@let@token.\else.\null\fi\xspace}
\makeatother

\crefname{section}{sec.}{sec.}
\Crefname{section}{Sec.}{Sec.}
\crefname{figure}{fig.}{fig.}
\Crefname{figure}{Fig.}{Fig.}

\definecolor{myred}{rgb}{0.600, 0.1, 0.1}
\definecolor{myblue}{rgb}{0.1,0.1, 0.600}

\usepackage{hyperref}
\hypersetup{
	colorlinks=true,
}
\threedvfinalcopy
\def\threedvPaperID{374} 

\ifthreedvfinal\fi

\title{Exemplar Fine-Tuning for 3D Human Model Fitting \\Towards In-the-Wild 3D Human Pose Estimation\thanks{Webpage
		: \url{https://github.com/facebookresearch/eft}}}

\author{Hanbyul Joo
	\hspace{0.3in} Natalia Neverova
	\hspace{0.3in} Andrea Vedaldi
	\vspace{5pt}
	\\
	{Facebook AI Research}
}

\begin{document}
	\maketitle
	\thispagestyle{empty}

	\begin{strip}
		\vspace{-3em}
		\captionsetup{type=figure} 
		\begin{center}
			\includegraphics[trim={0 0 0 430},clip,width=\linewidth]{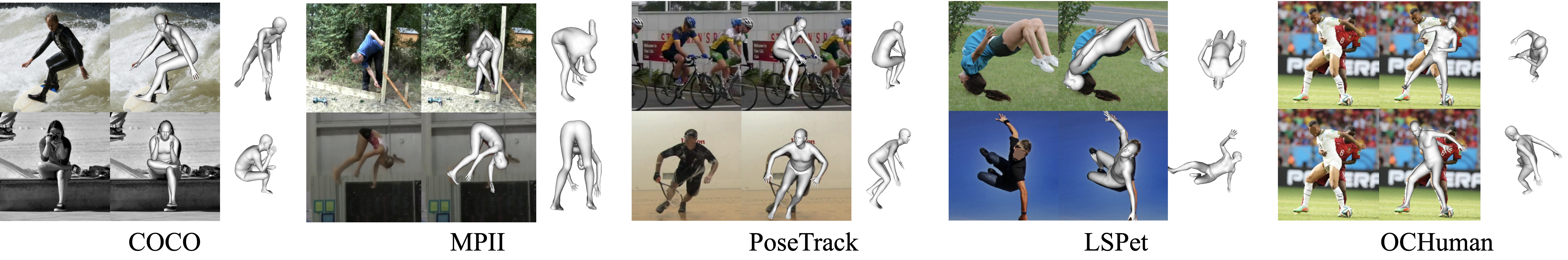}
			\captionof{figure}{Examples of 3D pseudo-ground truth annotations produced by our method, Exemplar Fine-Tuning (EFT), for images of people from COCO~\cite{lin2014microsoft}, MPII~\cite{Andriluka-14}, PoseTrack~\cite{PoseTrack}, LSPet~\cite{Johnson11}, and OCHuman~\cite{pose2seg2019} datasets. 
			}\label{fig:eft_teaser}
		\end{center}
	\end{strip}

	\section*{\hspace*{\fill}Abstract\hspace*{\fill}}\label{sec:abstract}
	
	\textit{Differently from 2D image datasets such as COCO, large-scale human datasets with 3D ground-truth annotations are very difficult to obtain in the wild.
		In this paper, we address this problem by \emph{augmenting} existing 2D datasets with high-quality 3D pose fits.
		Remarkably, the resulting annotations are sufficient to train from scratch 3D pose regressor networks that outperform the current state-of-the-art on in-the-wild benchmarks such as 3DPW\@.
		Additionally, training on our augmented data is straightforward as it does not require to mix multiple and incompatible 2D and 3D datasets or to use complicated network architectures and training procedures.
		This simplified pipeline affords additional improvements, including injecting extreme crop augmentations to better reconstruct highly truncated people, and incorporating auxiliary inputs to improve 3D pose estimation accuracy.
		It also reduces the dependency on 3D datasets such as H36M that have restrictive licenses.
		We also use our method to introduce new benchmarks for the study of real-world challenges such as occlusions, truncations, and rare body poses.
		In order to obtain such high quality 3D pseudo-annotations, inspired by progress in internal learning, we introduce Exemplar Fine-Tuning (EFT).
		EFT combines the re-projection accuracy of fitting methods like SMPLify with a 3D pose prior implicitly captured by a pre-trained 3D pose regressor network.
		We show that EFT produces 3D annotations that result in better downstream performance and are qualitatively preferable in an extensive human-based assessment. }
	
	\section{Introduction}\label{sec:intro}
	
	While 3D human pose estimation has progressed significantly in the past few years, the availability of suitable training datasets is still problematic.
	Obtaining 3D annotations for realistic benchmarks collected under everyday scenarios is difficult.
	Thus, most of the available 3D annotations are for datasets such as H36M~\cite{h36m_pami, ionescu2011latent} and MPI-INF-3DHP~\cite{mehta2017monocular}, captured mostly indoors or in laboratory conditions.
	In order to improve generalization, previous approaches propose relatively complicated training procedures such as adversarial learning~\cite{kanazawa2018end, humanMotionKanazawa19} or SMPLify-in-the-loop~\cite{kolotouros2019spin} that combine these 3D datasets with more realistic images with 2D keypoint annotations.
	For the same reason, using current 3D datasets for evaluation is also not necessarily representative of performance in the real world.
	Finally, many recent state-of-the-art approaches~\cite{kanazawa2018end,humanMotionKanazawa19,kolotouros2019spin,kocabas2019vibe} rely on SMPL fits on the H36M dataset~\cite{h36m_pami, ionescu2011latent} which, due to licensing restrictions, make reproducibility problematic.
	
	In this paper, we address these limitations by augmenting existing large-scale 2D datasets such as COCO~\cite{lin2014microsoft}, MPII~\cite{Andriluka-14},  PoseTrack~\cite{PoseTrack}, LSPet~\cite{Johnson11} with pseudo-ground truth (GT) 3D annotations, as shown in \cref{fig:eft_teaser}.
	While still inferred from single images, the quality of thus produces annotations is surprisingly high.
	To show this, we demonstrate that the new data (e.g., COCO with our pseudo annotations) is \emph{sufficient} to train state-of-the-art 3D pose regressor networks by itself, outperforming previous methods trained on the combination of datasets with 3D and 2D ground-truth~\cite{kanazawa2018end,sun2019human,kolotouros2019spin} on challenging benchmarks such as 3DPW\@.
	Notably, since the new dataset directly contains 3D annotations, which are unified across different benchmarks, training becomes straightforward without requiring complicated training techniques.
	Our datasets and codes are publicly available in our \href{https://github.com/facebookresearch/eft}{webpage}.
	
	Naturally, the success of our strategy depends on the quality of the pseudo-annotations that we can generate.
	Prior attempts at generating pseudo annotations for human pose such as UP-3D~\cite{Lassner:UP:2017} adopted established techniques such as SMPLify that fit the parameters of a 3D human model to the location of 2D keypoints, manually or automatically annotated in images~\cite{Bogo2016, joo2018, pavlakos19expressive}.
	However, they regularize the solution by means of a ``view-agnostic'' 3D pose prior, which incurs limitations:
	fitting ignores the RGB images themselves, and balancing between the 3D prior, learned separately in laboratory conditions, and the data term (\eg, 2D keypoint error) is difficult.
	In practice, we show that the quality of the resulting fits is insufficient for our purposes.
	
	To address these difficulties, we introduce \textbf{Exemplar Fine-Tuning} (EFT), an alternative pose fitting method inspired by recent progress in internal learning approaches~\cite{ulyanov18deep,zhang19an-internal,gadelha20deep}.
	The idea is to start from an image-based 3D pose regressor such as~\cite{Tung2017,kanazawa2018end,kolotouros2019spin}.
	The pretrained regressor \emph{implicitly} embodies an informal pose prior
	as a function $\bTheta = \Phi_\mathbf{w}(\mathbf{I})$, sending the image $\mathbf{I}$ to the parameters $\bTheta$ of the 3D body.
	This function can be interpreted as a \emph{conditional re-parameterization} of pose, where the conditioning factor is the image $\mathbf{I}$, and the new parameters are the weights $\mathbf{w}$ of the underlying neural network regressor.
	We can then fit 2D annotations similar to SMPLify, but using this as a new pose parameterization by fine-tuning the weights $\mathbf{w}$ rather than directly optimizing $\bTheta$.
	While similar techniques have been used in~\cite{Tung2017,alldieck2019learning,Zuffi:ICCV:2019} to improve test-time performance, they have not been rigorously assessed against classical optimization approaches such as SMPLify, nor they have been used to generate pseudo-labels.
	Our extensive quantitative and qualitative analysis demonstrate that EFT results in high quality fits simply by minimizing the 2D reprojection loss without the use of an explicit 3D pose prior.
	
	Using our 3D pseudo-annotations simplifies training 3D pose regressors and makes it easier to incorporate additional improvements, of which we explore two.
	First, we introduce extreme crop augmentation to train regressors that work better with truncated human bodies (e.g., upper body only)~\cite{Rockwell2020}.
	Second, we demonstrate that auxiliary inputs, such as color-coded segmentation maps~\cite{kirillov2019panoptic,kirillov2020pointrend} or DensePose IUV encodings~\cite{guler2018densepose}, can further improve the 3D human pose estimation accuracy, outperforming previous state-of-the-art approaches~\cite{kanazawa2018end,sun2019human,kolotouros2019spin,choi2020pose2mesh, moon2020i2l} simply by training on the COCO data.
	Compared to the previous approaches~\cite{Rockwell2020,rong2019delving, xu2019denserac, omran18neural}, these techniques are much easier to implement on top of our pseudo-GT dataset.
	
	We also show that our pseudo-annotations are useful to benchmark models, not just to train them.
	For this, we introduce a new 3D benchmark representative of real-world scenarios not captured by existing indoor or outdoor datasets such as 3DPW~\cite{vonMarcard2018}.
	We include:
	(1) COCO validation images~\cite{lin2014microsoft} representative of various in-the-wild scenes,
	(2) LSPet~\cite{Johnson11} images for gymnastic poses, and
	(3) OCHuman~\cite{pose2seg2019} for severe occlusions.
	We use EFT to generate pseudo-GT annotations and assess them via a large human study conducted on Amazon Mechanical Turk (AMT).
	Furthermore, we propose a new protocol for the 3DPW dataset to assess performance when only part of the human body is visible (truncation).
	Overall, the scenarios in the new 3D benchmark complement 3DPW\@; furthermore, they contain a far large number of different subjects, and thus much greater diversity.
	Assessed against this difficult data, our networks still outperform competitors.
	
	Summarizing, our contributions are:
	(1) providing large scale and high quality pseudo-GT 3D pose annotations that are sufficient to train state-of-the-art regressors without indoor 3D datasets;
	(2) running an extensive analysis of the quality of the pseudo-annotations generated via EFT against alternative approaches;
	(3) demonstrating the benefits of integrating the pseudo-annotations with extreme crop augmentation and auxiliary input representations;
	(4) introducing new 3D human pose benchmarks to assess regressors in less studied real-world scenarios.

	\section{Related Work}
	
	Deep learning has significantly advanced 2D pose recognition~\cite{cao2017realtime,cao2018openpose,Wei2016,Newell-16,xiao2018simple,sun2019deep}, facilitating the more challenging task of 3D reconstruction~\cite{tan17indirect,Tung2017,martinez2017simple,pavlakos2017coarse,pavlakos2018ordinal,omran18neural,pavlakos18learning,kanazawa2018end,kolotouros19convolutional,xiang2019monocular, Cheng_2019_ICCV, sun2019human, song2020human, ExPose:2020, moon2020i2l, choi2020pose2mesh}, which is our focus.
	
	\noindent \textbf{Single-view 3D Human Pose Estimation.}
	Single-view 3D pose reconstruction methods differ in how they incorporate a 3D pose prior and in how they perform the prediction.
	\emph{Fitting-based methods} assume a 3D body model such as SCAPE~\cite{anguelov2005scape}, SMPL~\cite{Loper2015}, SMPL-X\cite{pavlakos19expressive}, and STAR~\cite{STAR:2020}, and use an optimization algorithm to fit it to the 2D observations.
	While early approaches~\cite{guan09estimating,sigal08combined} required manual input, starting with SMPLify~\cite{Bogo2016} the process has been fully automatized, then improved in~\cite{Lassner:UP:2017} to use silhouette annotations, and eventually extended to multiple people~\cite{zanfir18monocular}. A recent method replaces the gradient descent update rule by a learned deep network~\cite{song2020human}.
	\emph{Regression-based methods}, on the other hand, predict 3D pose directly.
	The work of~\cite{Ramakrishna2012} uses sparse linear regression that incorporates a tractable but somewhat weak pose prior.
	Later approaches use instead deep neural networks, and differ mainly in the nature of their inputs and outputs~\cite{tan17indirect,Tung2017,martinez2017simple,pavlakos2017coarse,pavlakos2018ordinal,omran18neural,pavlakos18learning,kanazawa2018end,kolotouros19convolutional,xiang2019monocular,Cheng_2019_ICCV,xu2019denserac,kocabas2019vibe, moon2020i2l}.
	Some works start from a pre-detected 2D skeleton~\cite{martinez2017simple, choi2020pose2mesh} while others start from raw images~\cite{omran18neural, kanazawa2018end, kolotouros19convolutional, rong2019delving} or other proxy representation (e.g., DensePose outputs)~\cite{xu2019denserac}.
	Using a 2D skeleton relies on the quality of the underlying 2D keypoint detector and discards appearance details that could help fitting the 3D model to the image.
	Using raw images can potentially make use of this information, but training such models from current 3D indoor datasets might fail to generalize to unconstrained images.
	Hence several papers combine 3D indoor datasets with 2D in-the-wild ones~\cite{xiang2019monocular,kolotouros19convolutional,Tung2017,pavlakos18learning,omran18neural,kanazawa2018end,kocabas2019vibe,sun2019human,song2020human}.
	Methods also differ in their output, with some predicting 3D keypoints directly~\cite{martinez2017simple}, some predicting the parameters of a 3D human body model~\cite{kanazawa2018end,xiang2019monocular}, and others volumetric heatmaps for the body joints~\cite{pavlakos2017coarse} or meshes~\cite{kolotouros19convolutional, choi2020pose2mesh}.
	Finally, \emph{hybrid methods} such as SPIN~\cite{pavlakos19expressive} or MTC~\cite{xiang2019monocular} combine fitting and regression approaches.
	
	\noindent \textbf{3D Reconstruction Without Paired 3D Ground-truth.}
	Fitting methods such as SMPLify~\cite{xiao2018simple} only require 2D keypoint annotations and a human body model acquired independently on mocap data, but the output quality depends on the initialization, which is problematic for atypical poses.
	These methods can also use empirical pose priors by collecting a large number of samples in laboratory conditions~\cite{holden2015learning, akhter2015pose,pavlakos19expressive}, but those may lack realism.
	While most regression methods require paired 3D ground-truth, exceptions include~\cite{kanazawa2018end,zhou2017towards} that instead use 2D datasets and motion capture sequences to induce an empirical 3D prior enforced by means of adversarial learning~\cite{kanazawa2018end} or by using a geometric constraint on bone length ratios~\cite{zhou2017towards}.
	
	\noindent \textbf{Human Pose Datasets.}
	There are several in-the-wild datasets with sparse 2D pose annotations, including COCO~\cite{lin2014microsoft}, MPII~\cite{Andriluka-14}, Leeds Sports Pose Dataset (LSP)~\cite{Johnson10,Johnson11}, PennAction~\cite{zhang2013actemes}, and Posetrack~\cite{PoseTrack}.
	DensePose-COCO~\cite{guler2018densepose} offers dense surface point annotations instead.
	Annotating 3D human poses is much more challenging 
	as both specialized hardware and manual work are usually required.
	Examples include the Human3.6M dataset~\cite{h36m_pami}, Human Eva~\cite{sigal2010humaneva}, Panoptic Studio~\cite{joo2017panoptic}, and MPI-INF-3DHP~\cite{singleshotmultiperson2018}, which are hardly realistic.
	Two exceptions are 3DPW~\cite{vonMarcard2018} and PedX~\cite{kim2019pedx} that capture 3D ground-truth outdoors, but these are collected using specialized hardware (IMUs~\cite{vonMarcard2018} and LiDAR~\cite{kim2019pedx}) which limits data diversity.
	Previous attempts at creating 3D pseudo-annotations include UP-3D~\cite{Lassner:UP:2017}, using SMPLIfy, ExPose~\cite{choutas20monocular}, using SMPLIfy-X, and the work of Arnab \etal~\cite{arnab2019exploiting}, using a multi-frame optimization.
	Their use of traditional fitting methods limits the quality and scope of the pseudo annotations, due to the challenge in modeling generic human pose priors.
	
	
	

	\section{Exemplar Fine Tuning}\label{s:method}
	
	We are interested in 3D reconstruction of people depicted in images.
	Following~\cite{anguelov2005scape,Loper2015,joo2018,pavlakos19expressive, STAR:2020}, the human is represented by a parametric model that spans variation in both body shapes and motions.
	For example, the SMPL~\cite{Loper2015} model parameters $\bTheta = (\btheta, \bbeta)$ comprise the pose $\btheta \in \mathbb{R}^{24\times 3}$, which controls the rotations of 24 body joints, and the shape $\bbeta \in \mathbb{R}^{10}$, which controls body shapes by means of 10 principal directions of variations.
	The 3D locations $\mathbf{J}\in \mathbb{R}^{24\times3}$ of the body joints are obtained by first finding their configuration at rest using the shape parameters $\bbeta$, and then by applying the joint rotations  $\bTheta$
	\footnote{SMPL also includes a mesh component that deforms along with the skeleton, but we ignore it here as the major loss constrains only the 3D location $\mathbf{J}$ of the joints.}
	resulting in the \emph{posing function}
	$
	\mathbf{J} = M(\bTheta).
	$
	
	For reconstruction, \textbf{fitting-based approaches}~\cite{Bogo2016, Lassner:UP:2017} take 2D image cues such as joints, silhouettes, and part labels, and optimizes the model parameters $\bTheta$ to fit the 3D model to the 2D cues.
	For example, assume that the 2D locations $\bhj\in\mathbb{R}^{24\times 2}$ of the body joints are given.
	Furthermore, let $\pi : \mathbb{R}^3\rightarrow\mathbb{R}^2$ be the \emph{camera projection function}, mapping 3D points to their 2D image locations.
	Then, the optimal fit is given by:
	\begin{multline}
	(\bTheta^*, \pi^*)
	= \operatorname*{argmin}_{\bTheta, \pi}
	L_\text{2D}
	\left(
	\pi \left( M( \bTheta) \right), \bhj
	\right)
	+\\
	\gamma L_\text{pose}\left(\bTheta \right) 
	+\lambda L_\text{shape}\left(\bbeta \right),
	\label{eq:fitting_method}
	\end{multline}
	where the data term $L_\text{2D}$ is the re-projection error between reconstructed and observed 2D keypoints.
	Since the viewpoint of the image is unknown, the camera parameters $\pi$ must be optimized together with the body parameters $\bTheta$.
	The pose prior term $L_\text{pose}$ prioritizes plausible solutions, compensating for the fact that 2D keypoints do not contain sufficient information to infer 3D pose uniquely, and $\gamma$ is a scalar to adjust the weight of the prior term.
	SMPLify~\cite{Bogo2016} uses multiple prior terms including the pose prior expressed via a mixture of Gaussians or na\"{\i}ve thresholding, using a separate 3D motion capture dataset~\cite{cmu-graphics,akhter2015pose, AMASS:2019} to learn the prior parameters beforehand, as well as an angle prior to penalize unnatural bending.
	$L_\text{shape}$ is a shape prior term and $\lambda$ is its weight. 
	However, the success in optimizing~\eqref{eq:fitting_method} depends strongly on the quality of the heuristic used for initialization~\cite{Bogo2016,Lassner:UP:2017} (e.g.,~a multi-stage approach by first aligning the torso and then optimizing the limbs together) and balancing the weights between the data term and multiple prior terms is crucial.
	
	In contrast, \textbf{regression-based approaches} predict the model parameters $\bTheta$ directly from image-based cues $\bI$ such as raw RGB values~\cite{kanazawa2018end} and sparse~\cite{martinez2017simple} and dense~\cite{xu2019denserac} keypoints.
	The mapping is implemented by a neural network $\bTheta = \Phi_\mathbf{w}(\bI)$.
	Learning uses a number of example images $\bI_i$ with ground truth annotations for 2D keypoints $\bhj_i$, 3D keypoints $\hat{\mathbf{J}}_i$ and SMPL model parameters $\hat{\bTheta}_i$, which may in whole or in part be available, depending on the dataset (e.g.~\cite{h36m_pami,mehta2017monocular} for 3D and~\cite{lin2014microsoft,Andriluka-14} for 2D annotations).
	Learning then optimizes the following training objective:
	\begin{multline}\label{eq:learning_based_loss}
	\mathbf{w}^* =
	\operatorname*{argmin}_{\mathbf{w}}
	\frac{1}{N}
	\sum_{i=1}^N
	\Big[
	L_\text{2D}
	\left(
	\pi \left(M(\bphi_\mathbf{w}(\bI_i))\right), \bhj_i
	\right)\\
	+ \mu_i
	L_\text{3D}
	\left(
	M(\bphi_\mathbf{w}(\bI_i)), \hat{\mathbf{J}}_i
	\right)
	+ \tau_i
	L_{\bTheta}
	\left(
	\bphi_\mathbf{w}(\bI_i), \hat{\bTheta}_i
	\right)
	\Big],
	\end{multline}
	where $\mu_i$ and $\tau_i$ are loss-balancing coefficients which can be set to zero for samples that do not have 3D annotations.
	The camera projection $\pi$ is often predicted as additional output of the neural network $\bphi$~\cite{kanazawa2018end,humanMotionKanazawa19, xu2019denserac,kolotouros2019spin}.
	
	During training, the regressor $\bphi_{\mathbf{w}^*}(\bI)$ learns implicitly a prior on possible human poses, conditioned on the 2D input cues $\bI$.
	This is arguably stronger than the prior $L_\text{prior}$ used by fitting methods, which are learned separately, on different 3D data, and, being unconditional, tend to regress to a mean solution.
	On the other hand, fitting methods explicitly minimize the 2D re-projection error $L_\text{2D}$ at test time, thus resulting in better 2D fits than regression methods that only minimize this term during training.
	
	\textbf{Exemplar Fine-Tuning} (EFT) combines the advantages of fitting and regression methods.
	The idea is to interpret the network $\Phi$ as a re-parameterization 
	$
	\bTheta(\mathbf{w}) 
	= (\btheta(\mathbf{w}),\bbeta(\mathbf{w}))
	= \Phi_{\mathbf{w}}(\bI)
	$
	of the model as a function of the network parameters $\mathbf{w}$.
	With this, we can rewrite \cref{eq:fitting_method} as
	$
	\bTheta^* = \bTheta(\mathbf{w}^+)
	$
	where
	\begin{multline}\label{eq:exemplar_loss}
	\mathbf{w}^+
	=
	\operatorname*{argmin}_{\mathbf{w}}
	L_\text{2D}
	\left(
	\pi(M(\bTheta(\mathbf{w}))),
	\bhj 
	\right) \\
	+ \gamma \| \mathbf{w} - \mathbf{w}^* \|^2_2
	+\lambda L_\text{shape}\left(\bbeta \right),
	\end{multline}
	The second term requires the network parameters $\mathbf{w}$ to be close to the pre-trained regressor parameters $\mathbf{w}^*$.
	The shape regularizer $L_\text{shape} = \| \bbeta(\mathbf{w}) \|^2_2$
	favors the default SMPL shape parameters (as in \cref{eq:fitting_method}), but in practice its effect is very small ($\lambda = 0.001$, see \cref{table:postprocessing_3dpw}).
	
	Compared to \cref{eq:fitting_method}, we dropped the explicit pose prior $L_\text{pose}$, hypothesizing that a prior is implicitly captured by the pre-trained network, with the added advantage of accounting for the input image $\bI$.
	Concretely, consider using a large value of $\gamma$ for the prior in \cref{eq:fitting_method,eq:exemplar_loss}.
	Traditional fitting methods would then fall back to predicting the mean pose implied by the prior $L_\text{pose}$, ignoring the input.
	In contrast, in EFT a large value of $\gamma$ falls back to the `best guess' of the pre-trained regressor network for the pose of the observed input.
	Therefore, adjusting $\gamma$ is much easier in EFT\@, and, in supp.~material, we demonstrate that EFT is not sensitive to $\gamma$.
	In particular, fitting the pose prior in methods such as SMPLify may nullify the advantage of using a better regressor for initializing the pose.
	Instead, using better pose regressors in EFT generally result in better performance (see \cref{table:postprocessing_3dpw} and supp.~material).
	In our experiments, we employ EFT by setting $\gamma=0$ to produce our pseudo annotations, using instead early stopping for regularization, starting from $\mathbf{w} = \mathbf{w}^*$ (\cref{s:resultoverffing}).
	In practice, we use less than 100 EFT iterations for all fits.
	
	
	Note that, while~\cref{eq:learning_based_loss,eq:exemplar_loss} are optimized in a similar manner, the goals are very different:
	while~\cref{eq:learning_based_loss} is averaged over the training set and minimized to learn the parameters $\mathbf{w}^*$ of the model, \cref{eq:exemplar_loss} is optimized on a single example and only to find a better fit $\bTheta^* = \bTheta(\mathbf{w}^+)$ of the model to it.
	After an individual update $\bTheta^*$ is obtained, $\mathbf{w}^+$ is discarded.

	\noindent\textbf{Implementation details.}
	For the network $\Phi_{\mathbf{w}^*}$, we use HMR~\cite{kanazawa2018end} pre-trained using SPIN~\cite{kolotouros2019spin}.
	For robustness, we change the perspective projection model used in SPIN back to the weak-perspective projection of HMR~\cite{kanazawa2018end} and fine-tune the SPIN network to work correctly with this model (this step does not noticeably affect the performance of the model).
	The input RGB image $\bI$ is a crop, 224 $\times$ 224, around the 2D keypoint annotations.
	\cref{eq:exemplar_loss} is optimized using Adam~\cite{kingma2014adam} with the default PyTorch parameters and a learning rate\footnote{This learning rate is chosen as a smaller value than the learning rate $5 \times 10^{-5}$ used for training 3D pose regressor.} of $5 \times 10^{-6}$, which is sufficient to cover both good and bad initializations.  We switch off batch normalization and dropout.
	We iterate until the average 2D re-projection loss is less than 3 pixels, or up to a maximum of 50 iterations (100 for OCHuman~\cite{pose2seg2019} as the initial regressed pose tend to contain larger errors).
	%
	Although the input 2D keypoints are manually annotated, they still contain non-negligible errors. In particular, the heights of hips and the ankles are not consistently annotated, causing foreshortening (examples are shown in supp.~material).
	To compensate, the hip locations are ignored while optimizing~\cref{eq:exemplar_loss} and we add a loss term to match the orientation of the lower leg, encouraging the reconstruction of the orientation of the vector connecting the knee to the ankle. We use $\gamma=0$ in ~\cref{eq:exemplar_loss} unless otherwise mentioned.
	Finally, we discard samples as unreliable by checking the maximum value of the SMPL shape parameters and of the 2D keypoint loss, rejecting if these are larger than 5 and 0.01 respectively. 
	
	\section{The EFT Training and Validation Datasets}\label{s:benchmark}

	The main application of EFT is augmenting existing 2D human pose datasets with 3D pseudo-ground truth annotations for downstream model training and validation (benchmarking).
	We experiment with augmenting the COCO~\cite{lin2014microsoft}, MPII~\cite{Andriluka-14}, LSPet~\cite{Johnson11}, PoseTrack~\cite{PoseTrack}, and OCHuman~\cite{pose2seg2019} datasets.
	Most of these datasets come with a Train and Val split, which we use.\footnote{We make a random split for LSPet, and for OCHuman we use Val set as training and Test set as testing data.}
	We discard samples that fail the EFT sanity check discussed in~\cref{s:method}.
	For COCO-Train and COCO-Val, we only retain samples with at least 6 keypoints annotations.
	We also consider a subset of COCO-Train, named ``COCO-Part'', used by~\cite{kolotouros19convolutional}, which more conservatively only uses instances with 12 keypoint annotations for which all limbs are present.
	For COCO-Val, LSPet and OCHuman we further carry out a manual filtering step, described below.
	We use the notation $[\cdot]_{\text{EFT}}$ to denote an EFT-augmented version of a dataset and thus obtain the $[\text{COCO-Train}]_{\text{EFT}}$, $[\text{COCO-Part}]_{\text{EFT}}$, $[\text{MPII}]_{\text{EFT}}$, $[\text{LSP}]_{\text{EFT}}$, $[\text{PoseTrack}]_{\text{EFT}}$, and $[\text{OCHuman}]_{\text{EFT}}$ datasets, summarized in~\cref{table:eft_dataset}.
	All datasets are publicly available in our webpage.
	
	\noindent \textbf{Human Study and Validation.}
	We conduct a human study and validation on Amazon Mechanical Turk (AMT) to assess the quality of the EFT fits.
	First, we use A/B testing to compare EFT and SMPLify fits.
	To this end, we show 500 randomly-chosen images from the MPII, COCO and LSPet datasets to human annotators in AMT and ask them whether they prefer the EFT or the SMPLify reconstruction.
	Each sample is evaluated by three different annotators, showing the input image and two views of the 3D reconstructions, from the same viewpoint as the image and from the side.
	Examples are shown in supp.~material. 
	Our method was preferred \textbf{61.8\%} of the times with a majority of at least 2 votes out of 3, and obtained \textbf{59.6\%} favorable votes by considering the 1500 votes independently.
	We found that in many cases the perceptual difference between SMPLify and EFT is subtle, but SMPLify suffers more from bad initialization due to occlusions and challenging body poses.
	
	We further conduct a full human assessment of COCO-Val, LSPet and OCHuman to validate the annotations before using them for benchmarking purposes.
	To this end, we show each sample to three annotators at random, asking whether the estimated 3D annotation accurately describes the pose of the target subject in the scene or not.
	Conservatively, we accept a sample only if all three annotators agree to accept it.
	The final number of samples and rejection rates are shown in \cref{table:eft_dataset}.
	The rejection rates ($\approx 20\%$) are relatively high due to the strict selection criteria (the rate would be only $\approx 4\%$ if were to accept samples with only two `accept' votes).

	\begin{table}[t]
		\centering
		\rowcolors{1}{}{lightgray}
		\footnotesize
		\begin{tabular}{l c c c c c}
			\toprule
			\textbf{Dataset} & \!\!\textbf{\# Initial}\! &  \multicolumn{2}{c}{\textbf{\# Final}} & \!\!\textbf{Human}\!\! &
			\textbf{Rej. Rate} \\
			&                                & Train & Val  &      & \\
			\midrule
			COCO-Val~\cite{lin2014microsoft} & 13K   &      & 10K  & \cmark & 20.5\% \\
			LSPet~\cite{Johnson11}           & 8.3K  & 3.3K & 2.2K & \cmark & 18.6\% \\
			OCHuman~\cite{pose2seg2019}      & 10K  & 2.6K & 1.7K & \cmark & 24.1\% \\
			\midrule
			COCO-Train                       &  79K     & 79K  & ---  & \xmark & --- \\
			COCO-Part                        &  28K     & 28K  & ---  & \xmark & --- \\
			MPII                             &  15K     & 14K  & ---  & \xmark & --- \\
			PoseTrack-Train                        &   29K   & 22K  & ---  & \xmark & --- \\
			\bottomrule
		\end{tabular}
		\caption{
			\textbf{Summary of EFT validation (top) and training (bottom) datasets.} The `Human' and `Rej. Rate' columns show whether manual filtering is used with the rejection rates.}\label{table:eft_dataset}
	\end{table}
	
	\begin{figure}
		\includegraphics[trim={0 4cm 4cm 0},clip,width=\linewidth]{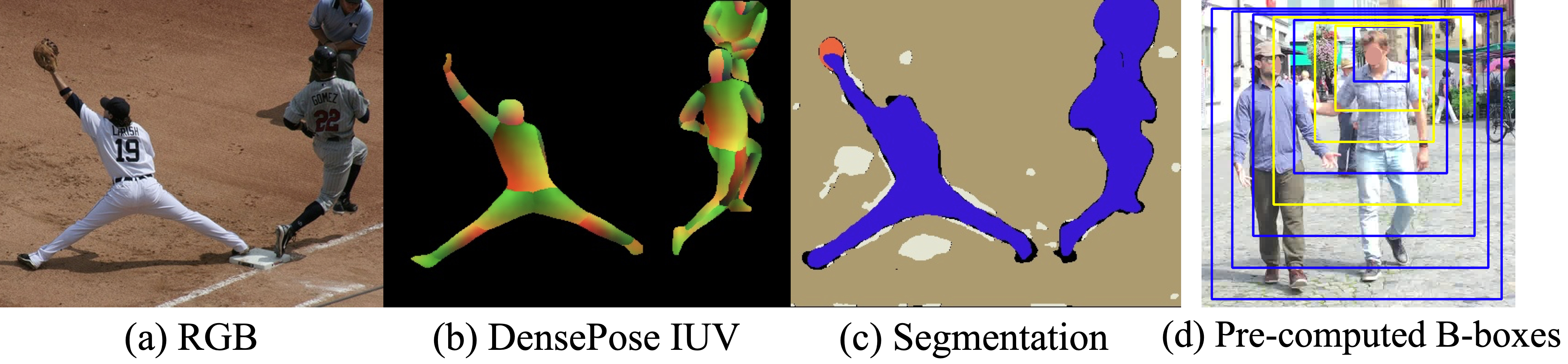}
		{\footnotesize \hspace*{0.1cm} (a) RGB \hspace*{0.4cm} (b) DensePose IUV \hspace*{0.1cm} (c) Segmentation \hspace*{0.3cm} (d) Boxes}\vspace*{2pt}
		\caption{(a) An example RGB image, and auxiliary inputs from (b) DensePose~\cite{guler2018densepose} and (c) color-coded segmentation maps~\cite{wu2019detectron2, kirillov2019panoptic} (c).  (d) Pre-computing different levels of bounding boxes for crop augmentation and robustness test on truncated body. The yellow boxes (levels 1,2, and 4) are used for evaluation in \cref{table:upperbody_3dpw}.}\label{fig:eft_bbox}
		
	\end{figure}
	
	\noindent \textbf{Truncated 3DPW Dataset.}\label{s:benchmark_truncated_3dpw}
	In order to assess the robustness of algorithms to people only partially visible due to view truncation (a very common case in applications as similarly addressed in~\cite{Rockwell2020}), we also propose a new protocol for the 3DPW dataset using a pre-defined set of aggressive image crops (see~\cref{fig:eft_bbox}).
	As shown in \cref{fig:eft_bbox} (d), crops are generated by computing bounding boxes in between the full bounding box (level 7) of a person and a much smaller bounding box containing only his/her face (level 0) whose 2D locations is computed by projecting the SMPL fits on images.
	For each human instance, we generate eight intermediate bounding boxes in this manner (levels 0 to 7).

	\section{Training Pose Regressors with EFT datasets}\label{s:details}
	
	Our EFT datasets allow us to train 3D pose regressors from scratch using a simple and clean pipeline.
	Specifically, we train the HMR model~\cite{kanazawa2018end} (without the discriminator) using the same hyper parameters as used in SPIN~\cite{kolotouros2019spin}. 
	%
	A key advantage in using the pseudo-GT annotations is that the training pipeline becomes straightforward, providing opportunities to apply additional techniques to improve performance.
	We explore two such techniques: applying extreme crop augmentation and inputting auxiliary inputs.
	
	\noindent \textbf{Augmentation by extreme cropping.}
	A shortcoming of previous 3D pose estimation methods is that it is assumed that most of the human body is visible in the input image~\cite{xu2019denserac,guler2019holopose,kanazawa2018end,kolotouros2019spin}.
	However, humans captured in real-world videos are often truncated, so that only the upper body or even just the face is visible, dramatically increasing the ambiguity of pose reconstruction.
	In order to train models that are more robust to truncation, we propose to augment the training data with extreme cropping.
	While the same challenge has been recently addressed by~\cite{Rockwell2020}, our approach is more straightforward because we already have full 3D annotations --- we only need to randomly crop the training images.
	In practice, we generate random crops in the same way as described in~\cref{s:benchmark} for the Truncated 3DPW dataset.
	During the training, we trigger crop augmentation with 30\% chance, randomly choosing a truncated bounding box among the pre-computed ones shown in \cref{fig:eft_bbox} (d).
	
	\noindent \textbf{Auxiliary Input for 3D Pose Regressor.}
	Recent methods demonstrate that other types of input encoding such as DensePose or body part segmentation can improve 3D pose regressors~\cite{guler2018densepose, xu2019denserac, rong2019delving, alldieck2019learning}.
	Training with such auxiliary inputs is also straightforward in our pipeline.
	To show this, we train a pose regressor by concatenating the standard RGB input with an additional input encoding.
	We test color-coded segmentation maps~\cite{kirillov2019panoptic, kirillov2020pointrend} and DensePose map, by pre-computing these representations for all images in the datasets using Detectron2~\cite{wu2019detectron2}, as shown in \cref{fig:eft_bbox} (b,c). To accept 6 channel inputs (RGB concatenated with a color-coded auxiliary input), we modify the first layer of HMR (i.e. the first layer of ResNet50) by duplicating the initial weights.

	
	\section{Results}\label{s:results}
	
	Our key result is showing that the quality of the EFT-generated pseudo GT is sufficient to train from scratch state-of-the-art 3D pose regressors, outperforming previous methods (\cref{s:eft-train,s:learning-auxiliary}) in the in-the-wild benchmark.
	We also use EFT to generate benchmark representative of difficult scenarios of practical importance and run a large comparison of state-of-the-art regressors using it (\cref{s:eft-bench,s:benchmark-trunc3dpw}).
	Finally, we use EFT as a post-processing step with off-the-shelf regressors (\cref{s:eft-post}), and provide additional analysis in (\cref{s:resultoverffing}).
	
	\subsection{Datasets}\label{s:datasets}
	
	In addition to the EFT datasets of~\cref{s:benchmark}, we also test the datasets with ground-truth 3D pose annotations, including H36M~\cite{h36m_pami, ionescu2011latent}, MPI-INF-3DHP~\cite{mehta2017monocular}, collected in laboratory conditions.
	For these datasets, we use the existing SMPL fittings on H36M~\cite{kanazawa2018end} and MPI-INF-3DHP~\cite{kanazawa2018end, kolotouros2019spin}.
	We use the 3DPW test set~\cite{vonMarcard2018} as the major target benchmark that is captured outdoor and comes with 3D ground truth obtained by using IMUs and cameras.
	
	\begin{table}[t]
		\centering
		\rowcolors{1}{}{lightgray}
		\footnotesize
		\begin{tabular}{l l l}
			\toprule
			\textbf{Previous method}                         & \textbf{3DPW} $\downarrow$ & \textbf{H36M} $\downarrow$ \\
			\midrule
			HMR~\cite{kanazawa2018end}                       & 81.3                       & 56.8 \\
			DSD~\cite{sun2019human}                & 75.0                        & 44.3\\
			SPIN~\cite{kolotouros2019spin}                  & 59.2                       & 41.1\\
			I2L~\cite{moon2020i2l}                   & 57.7                       & 41.1\\
			Pose2Mesh~\cite{choi2020pose2mesh}                   & 58.3                       & 46.3 \\
			ExPose~\cite{ExPose:2020} (by SMPLify-X fits)                  & 60.7                       & ---\\
			Song~\etal~\cite{song2020human} (w/ 2D keypoints)                   & 55.9                       & 56.4\\
			VIBE (temporal)~\cite{kocabas2019vibe}            & 56.5                       & 41.5\\
			\midrule
			\multicolumn{3}{l}{\textbf{Direct 3D supervision, 3D pseudo-ground truth from EFT}}\\
			\midrule
			H36M  & 146.2   & 53.3\\
			MPI-INF-3DHP (MI)   & 127.8 & 110.6 \\
			3DPW-Train    & 91.0  & 131.4\\
			\midrule
			$[\text{LSP}]_{\text{EFT}}$ & 84.9 & 87.3 \\
			$[\text{MPII}]_{\text{EFT}}$    & 69.1  & 78.9 \\
			$[\text{PoseTrack}]_{\text{EFT}}$    & 72.6  & 88.6 \\
			$[\text{COCO-Part}]_{\text{EFT}}$   & \textbf{59.0}  & 66.9 \\
			$[\text{COCO-Train}]_{\text{EFT}}$    & \textbf{57.5} & 64.2 \\
			$[\text{COCO-Train}]_{\text{EFT}}$ + $[\text{PoseTrack}]_{\text{EFT}}$    & \textbf{56.5} & 63.5 \\
			$[\text{COCO-Part}]_{\text{EFT}}$ + H36M & \textbf{57.8} & 45.0 \\
			$[\text{COCO-Train}]_{\text{EFT}}$ + H36M & \textbf{55.5}  & 46.2 \\
			$[\text{COCO-Train}]_{\text{EFT}}$ + H36M + MI    & \textbf{54.7}  & 44.9 \\
			$[\text{COCO-Train}]_{\text{EFT}}$ + H36M + MI + 3DPW-T & \textbf{51.6}  & 44.0 \\
			\midrule
			$[\text{Co-Tr+PT+LSP}]^{ca}_{\text{EFT}}$ + H36M + MI    & \textbf{53.6}  & 45.6 \\
			$[\text{Co-Tr+PT+LSP+OC}]^{ca}_{\text{EFT}}$ + H36M + MI    & \textbf{54.6}  & 45.4 \\
			\midrule
			$[\text{COCO-Part}]_{\text{SPIN~\cite{kolotouros2019spin}}}$    & 70.9  & 78.2\\
			$[\text{COCO-Part}]_{\text{SMPLify}}$   & 72.7 & 81.0 \\
			\midrule
			$[\text{COCO-Part}]_{\text{EFT}}$ w/ Densepose   & \textbf{57.2}  & 64.0\\
			$[\text{COCO-Train}]_{\text{EFT}}$ w/ Densepose  & \textbf{56.1} & 64.7 \\
			
			\bottomrule
		\end{tabular}
		\caption{
			\textbf{Quantitative evaluation on 3DPW (PA-MPJPE) and H36M (protocol-2 using frontal view, PA-MPJPE) in $mm$.}
			Top: previous methods. Bottom: training using straight 3D supervision with actual and pseudo 3D annotations.
			We also compare generating pseudo-ground truth annotations using EFT, SPIN~\cite{kolotouros2019spin}, and SMPLify~\cite{Bogo2016, ExPose:2020}.
			We report in bold regression results that are better than the previously-reported state of the art (SPIN) that is based on the same network architecture. 
			\label{table:quant_public_db}
		}
	\end{table}

	\begin{table}[t]
		\centering
		\rowcolors{1}{}{lightgray}
		\scriptsize
		\begin{tabular}{lcc}
			\toprule
			\textbf{Methods}                         & \textbf{OCHuman} $\downarrow$ & \textbf{LSP} $\downarrow$ \\
			\midrule
			SPIN~\cite{kolotouros2019spin}                   & 104.1                      & --- \\
			\midrule
			H36M  & 165.4   & 201.8\\
			MPI-INF-3DHP (MI)   &158.1 & 190.2 \\
			3DPW (Train)    & 141.7  & 182.4\\
			\midrule
			$[\text{LSP}]_{\text{E}}$   & 129.7 & 105.7 \\
			$[\text{MPII}]_{\text{E}}$    & 113.3  & 127.0 \\
			$[\text{PT}]_{\text{E}}$    & 124.8  & 154.7 \\
			$[\text{CO-Pa}]_{\text{E}}$   & 104.8  & 108.9 \\
			$[\text{CO-Tr}]_{\text{E}}$    & 104.0 & 103.9 \\
			$[\text{CO-Pa}]_{\text{E}}$ + H36m & 103.2 & 106.1 \\
			$[\text{CO-Tr}]_{\text{E}}$ + H36m + MI + $[\text{PT}]_{\text{E}}$ & 103.1  & 99.9 \\
			$[\text{CO-Tr}]_{\text{E}}$ + H36m + MI + $[\text{PT}]_{\text{E}}$ + $[\text{LSP}]_{\text{E}}$ & 103.9  & 79.4 \\
			$[\text{CO-Tr}]_{\text{E}}$ + H36m + MI + $[\text{PT}]_{\text{E}}$ + $[\text{LSP}]_{\text{E}}$ + $[\text{OCH}]_{\text{E}}$\!\!\!\!\!\! & \textbf{92.7}  & \textbf{77.6} \\
			\bottomrule
		\end{tabular}
		\caption{
			\textbf{Quantitative evaluation on OCHuman and LSP datasets with our pseudo annotations}, in PA-MPJPE errors (mm).
		}\label{table:quant_new_benchmarks}
	\end{table}

	\subsection{EFT Datasets for Learning Models}\label{s:eft-train}
	
	In~\cref{s:benchmark} we have assessed the EFT-augmented datasets via a human study.
	We now assess them based on how well they work when used to train pose regressors from scratch\footnote{ResNet50 of HMR is initialized with ImageNet pretrained weights.}.
	To this end, we follow the procedure described in~\cref{s:details} and train the HMR model using the EFT datasets in isolation as well as in combination with other 3D datasets such as H36M.
	In all cases, we use as single training loss the prediction error against 3D annotations (actual or pseudo), with a major simplification compared to approaches that mix, and thus need to balance, 2D and 3D supervision.
	
	The results are summarized in \cref{table:quant_public_db}.
	The table (bottom) evaluates the regressor trained from scratch using straight 3D supervision on standard 3D datasets (H36M, MPI-INF-3DHP, 3DPW), the EFT-lifted datasets $[\text{MPII}]_{\text{EFT}}$, $[\text{LSP}]_{\text{EFT}}$, $[\text{COCO-Part}]_{\text{EFT}}$ and $[\text{COCO-Train}]_{\text{EFT}}$, as well as various combinations.
	Following~\cite{kanazawa2018end, kolotouros2019spin}, performance is measured in terms of reconstruction errors (PA-MPJPE) in $mm$ after rigid alignment on two public benchmarks: 3DPW and H3.6M\footnote{See the supp.~material for the result on the MPI-INF-3DHP benchmark.}.
	The models trained with indoor 3D pose datasets (H36M and MPI-INF-3DHP) perform poorly on the 3DPW dataset, which is collected outdoors.
	By comparison, training exclusively on the EFT datasets performs much better.
	Notably, the model trained \emph{only} on  $[\text{COCO-Part}]_{\text{EFT}}$ is already comparable to the previous state-of-the art method, SPIN~\cite{kolotouros19convolutional}, on this benchmark. Combining multiple training datasets improves performance further; the model trained with $[\text{COCO-Train}]_{\text{EFT}}$ outperforms SPIN achieving (57.5 $mm$) and the model trained with $[\text{COCO-Train}]_{\text{EFT}}$ + $[\text{PoseTrack}]_{\text{EFT}}$ is comparable to VIBE~\cite{kocabas2019vibe} that is the previous best method using a video input (using temporal cues).
	Combining both EFT datasets and 3D datasets shows even better results with the result (54.7 $mm$) achieved by the $[\text{COCO-Train}]_{\text{EFT}}$ + H36M + MPI-INF-3DHP combination, and the best one (51.6 $mm$) by including the 3DPW training data.
	
	We also compare EFT to SMPLify~\cite{Bogo2016} for generating the pseudo-annotations.
	In $[\text{COCO-Part}]_{\text{SMPLify}}$ we use the SMPLify output to post-process the output of the fully trained SPIN model, and in $[\text{COCO-Part}]_{\text{SPIN}}$ we use the fitting outputs that SPIN generates during learning~\cite{kolotouros2019spin}, which are publicly available.
	We include the result of ExPose~\cite{ExPose:2020} in~\cref{table:quant_public_db} that similarly built a pseudo-gt dataset using SMPLify-X with a manual filtering process by human annotators.
	As shown, the models trained on the EFT datasets perform better than the ones trained on the SMPLify-based ones, suggesting that the quality of the pseudo GT generated by EFT is better.
	
	\begin{table}[t]
		\centering
		\caption{
			\textbf{Effect of crop-augmentation on 3DPW (PA-MPJPE in $mm$) with different level of truncated inputs.} 
		}\label{table:upperbody_3dpw}
		\rowcolors{1}{}{lightgray}
		\centering
		\footnotesize
		\begin{tabular}{lcccc}
			\toprule
			\textbf{Training DBs}	&  \!\!\textbf{crop lev.1}\!\!\!\!\!	&  \!\textbf{lev.2}\! &  \!\textbf{lev.4}\! &  \textbf{orig.}\\
			\midrule
			HMR~\cite{kanazawa2018end} & 153.6 & 122.2 & 82.0 & 81.3 \\ 
			Rockwell \etal~\cite{Rockwell2020} & 133.2 & 106.1 & 78.0 & 71.9 \\
			SPIN~\cite{kolotouros2019spin} & 184.3 & 198.8 & 79.4 & 59.2\\
			VIBE~\cite{kolotouros2019spin} & 150.5 &  158.9 & 75.1& 56.5\\
			\midrule
			$[\text{CO-P}]_{\text{E}}$     & 168.7  & 145.0 & 69.7 & 59.0\\ 
			$[\text{CO-T}]_{\text{E}}$      & 113.5  & 97.4 & 66.3 & 57.5\\
			$[\text{CO-T}]_{\text{E}}$ + $[\text{PT}]_{\text{E}}$  &  118.0 &  100.1 & 68.3 & 56.5\\
			$[\text{CO-P}]^{ca}_{\text{E}}$     & 89.4  & 76.3 & 64.4 & 61.5\\
			$[\text{CO-T}]^{ca}_{\text{E}}$      & 91.5  & 76.4 & 63.4 & 59.2\\
			$[\text{CO-T}]^{ca}_{\text{E}}$ + $[\text{PT}]^{ca}_{\text{E}}$ \!\!\!\!\! &  88.5 &  75.9 & 63.4 & 59.3\\
			
			$[\text{CO+PT+LSP}]^{ca}_{\text{E}}$ + H36m + MI   \!\!\!\!\!\!\!  & 89.1  & 74.8 &\textbf{59.6} & \textbf{53.6} \\
			$[\text{CO+PT+LSP+OC}]^{ca}_{\text{E}}$ + H36m + MI   \!\!\!\!\!\!\!\!\!\!  & \textbf{84.5}  & \textbf{72.9} & 60.3 & 54.6 \\
			\bottomrule
		\end{tabular}
	\end{table}
	
	
	
	
	This result shows the significance of the indoor-outdoor domain gap and highlights the importance of developing in-the-wild 3D datasets, as these are usually closer to applications, thus motivating our approach. As it might be expected, networks trained exclusively on indoor dataset with GT 3D annotations (H36M, MPI-INF-3DHP, and 3DPW-train) show poor performance on 3DPW.
	Similarly, the networks exclusively trained on the the EFT-based datasets, which are `in the wild', are not as good when tested on H36M.
	However, the error decreases markedly once the networks are also trained using the H36M data (H36M, $[\text{COCO-Part}]_{\text{EFT}}$+H36M, and $[\text{COCO-Train}]_{\text{EFT}}$ + H36M +MPI-INF-3DHP).


	\subsection{Learning Models with Auxiliary Inputs}\label{s:learning-auxiliary}
	
	Next, we test the performance of models trained with RGB and auxiliary inputs.
	In \cref{table:quant_public_db} bottom, $[\text{COCO-Part}]_{\text{EFT}}$ w/ DensePose and $[\text{COCO-Train}]_{\text{EFT}}$ w/ DensePose show results obtained by using the concatenation of RGB and  DensePose IUV map as input. These additional inputs improve the accuracy on in-the-wild scenes, previously shown by~\cite{rong2019delving}.
	In particular, the models trained with $[\text{COCO-Train}]_{\text{EFT}}$ \textit{only} outperform the state-of-the-art \emph{video-based regressor} VIBE (56.1 vs 56.5 $mm$).
	In the sup.~mat.~we show that, instead, the segmentation encodings do not noticeably improve performance.

	\subsection{New 3D Human Pose Benchmarks}\label{s:eft-bench}
	
	\paragraph{EFT Datasets.} We evaluate the performance of various models on our new benchmark datasets with pseudo GT, OCHuman~\cite{pose2seg2019} and LSPet~\cite{Johnson11}.
	These datasets have challenging body poses, camera viewpoints, and occlusions.
	We found that most models trained on other datasets are struggling in these benchmarks, showing more than 100 $mm$ errors, as shown in table~\ref{table:quant_new_benchmarks}.
	The models trained with pseudo annotations on similar data (training sets of LSPet and OCHuman) show better performance (less than 100 $mm$).
	\vspace{-10pt}
	
	\paragraph{Testing with Truncated Input on 3DPW.}\label{s:benchmark-trunc3dpw}
	%
	We use the protocol defined in \cref{s:benchmark_truncated_3dpw} on 3DPW to assess performance on truncated body inputs. Among 8 different bounding box levels, we consider levels 1,2,4 that are shown in yellow in \cref{fig:eft_bbox} (Right).
	We use PA-MPJPE as in the original benchmark test.
	The result is shown in \cref{table:upperbody_3dpw}, where $[\cdot]^{ca}_{\text{E}}$ are the datasets where we apply crop augmentations.
	For comparison, we include a recent work of Rockwell \etal~\cite{Rockwell2020} that is trained to handle similar partial view scenes.
	
	While HMR, SPIN, VIBE, and our network trained with $[\text{COCO-Part}]_{\text{EFT}}$ without crop augmentation work poorly, we found that our models trained with $[\text{COCO-Train}]_{\text{EFT}}$ show better performance even without crop augmentation, even outperforming the work of Rockwell \etal~\cite{Rockwell2020}. This is because $[\text{COCO-Train}]_{\text{EFT}}$ already includes many such samples with severe occlusions. Note that $[\text{COCO-Train}]_{\text{EFT}}$ includes all samples with 6 or more valid 2D keypoint annotations.
	Our models trained with crop augmentation show much better performance even for scenes with extreme truncation (crop level 1 and 2).
	Note that crop augmentation does not degrade the performance of our models in the original 3DPW benchmark, as shown in \cref{table:quant_public_db}.
	\begin{table}[t]
		\centering
		\scriptsize
		\begin{tabular}{l|>{\columncolor{lightgray!70}}c|ccc|cc}
			\toprule
			\multirow{2}{*}{\textbf{Datasets}}	&   & \multicolumn{3}{c|}{\textbf{SMPLify}} & \multicolumn{2}{c}{\textbf{EFT}} \\
			& \multirow{-2}{*}{\textbf{No post}} & \!\!\textbf{M+D+P}\!\!\!\! & \textbf{D+P} & \textbf{D} &  \!\textbf{$\gamma=0$}\! &  \!\!\!\textbf{$\gamma, \lambda=0$}\!\!\!\!\\
			\midrule
			H36M  & 146.2 & \textcolor{myblue}{141.9} & \textcolor{myblue}{139.6} & \textcolor{myred}{152.5}   & \textcolor{myblue}{108.3}  & \textcolor{myblue}{108.2}\\
			$[\text{LSP}]_{\text{E}}$  & 84.9  & \textcolor{myblue}{77.6} & \textcolor{myblue}{69.7} & \textcolor{myblue}{75.4}  & \textcolor{myblue}{69.4} & \textcolor{myblue}{69.4} \\
			$[\text{MPII}]_{\text{E}}$ & 69.1 & \textcolor{myblue}{68.0} & \textcolor{myblue}{64.6} & \textcolor{myblue}{67.3} &  \textcolor{myblue}{63.4} & \textcolor{myblue}{63.4}\\
			CO+H3+M & 54.7  & \textcolor{myred}{61.5} & \textcolor{myred}{56.7} & \textcolor{myred}{57.2}           & \textcolor{myblue}{53.4} & \textcolor{myblue}{53.4}  \\
			CO+H3+M+W\! & 51.6  & \textcolor{myred}{60.2} & \textcolor{myred}{54.1}& \textcolor{myred}{52.0}        & \textcolor{myblue}{50.4} & \textcolor{myblue}{50.4}\\
			\midrule
		\end{tabular}
		\caption{\textbf{Quantitative evaluation on 3DPW by using SMPLify and EFT as post-processing}, in PA-MPJPE errors ($mm$). The second column shows the errors after direct regressions, and the next columns show the post-processing results. In SMPLify, M, D, P stand for multi-stage approach, data-term, and prior term respectively.
			The numbers in blue indicate the cases in which post-processing improves over the original predictions, and the red color indicates loss in performance. `CO+H3+M+W' stands for `$[\text{COCO-Train}]_{\text{EFT}}$ + H36M + $[\text{MI}]_{\text{EFT}}$ + 3DPW-T'.
			\label{table:postprocessing_3dpw}}
	\end{table}
	\subsection{EFT as A Post-processing Method}\label{s:eft-post}
	EFT can also be be used as a drop-in post-processing step on top of any given pose regressor $\Phi$ to improve its test-time performance.
	We compare EFT post-processing against performing the same refinement using a traditional fitting method such as SMPLify, starting from the same regressor for initialization.
	In this experiment, we use the `gold-standard setting' for each algorithm: for EFT, this is the same setup as in the other experiments, and for SMPLify, we use the best hyper-parameter setting determined in SPIN\@. See supp.~mat. for details.
	In the sup.~mat.~we also report results when the settings are exactly the same, except for the fitting method, and arrive to analogous conclusions.
	
	In \cref{table:postprocessing_3dpw} we carry out a quantitative evaluation on the 3DPW dataset.
	For initialization, we use various regressors $\Phi$ pre-trained on a number of different datasets, providing different initialization states.
	We then use EFT and SMPLify post-processing to fit the same set of 2D keypoint annotations, obtained automatically by means of the OpenPose detector~\cite{cao2018openpose}.
	For SMPLify, we performed an ablation study by turning off multi-stage trick (M) and prior terms (P), compared to the standard setting (M+D+P) used in SPIN and the original SMPLify literature~\cite{Bogo2016}.
	Similarly, we tested EFT performance without shape regularizer (i.e., $\lambda=0$).
	The key observation is that the performance of SMPLify post-processing depends on the initialization quality, the use of their multi-stage trick, and balancing between data and prior terms.
	Specifically, if the initialization is already good, as in the last row of \cref{table:postprocessing_3dpw}, SMPLify degrades the performance, particularly when the pose prior is used.
	This illustrates the difficulty in balancing data and prior terms, which may be difficult to do in practice. 
	In contrast, EFT does not suffer from such issues, improving the accuracy in \emph{all} cases, regardless of the quality of the initialization.
	This is also true when the shape regularizer is removed entirely.
	
	\begin{figure}
		\includegraphics[trim=100 35 100 100,clip,width=0.495\linewidth]{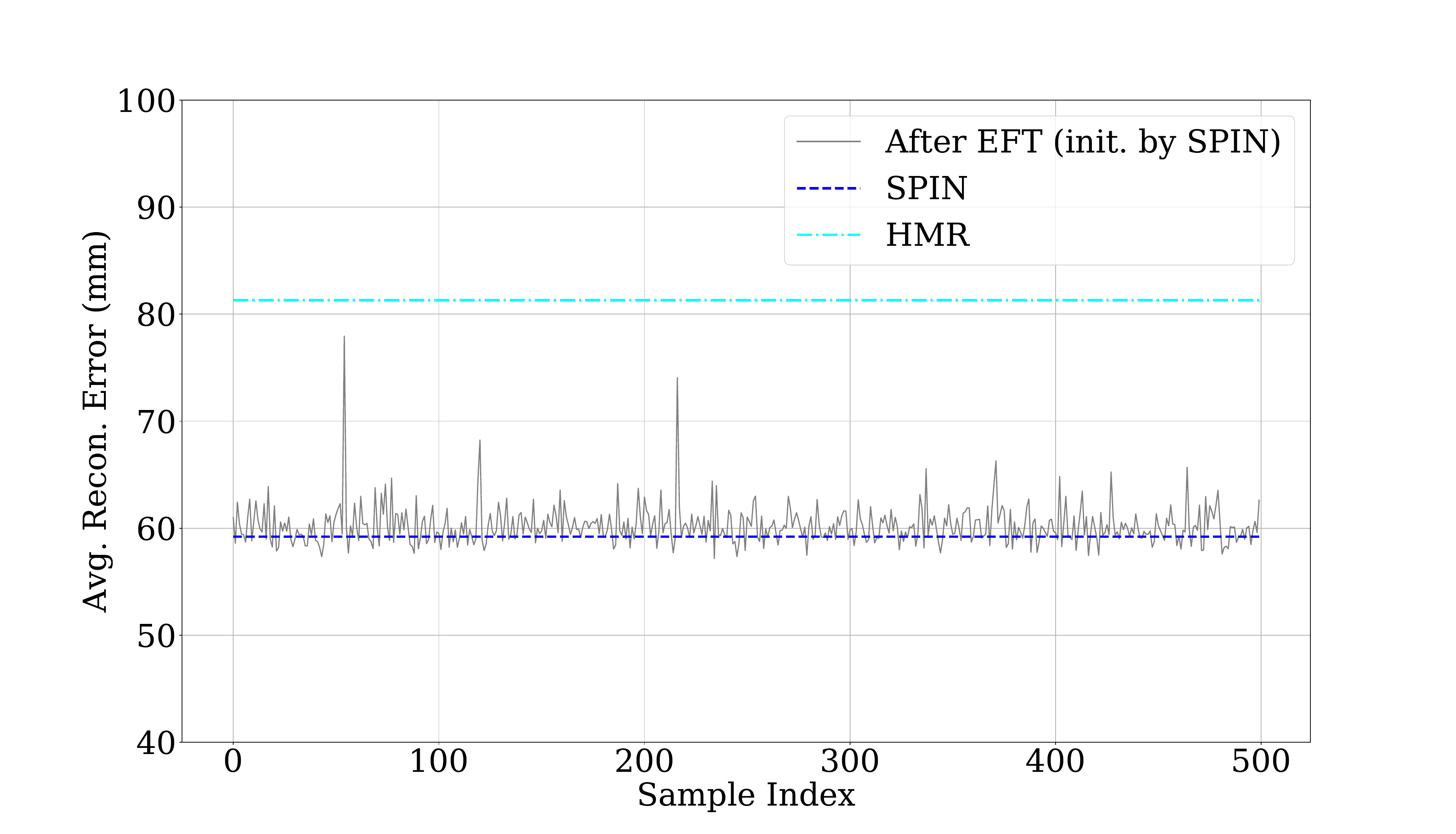}
		\includegraphics[trim=100 35 100 100,clip,width=0.495
		\linewidth]{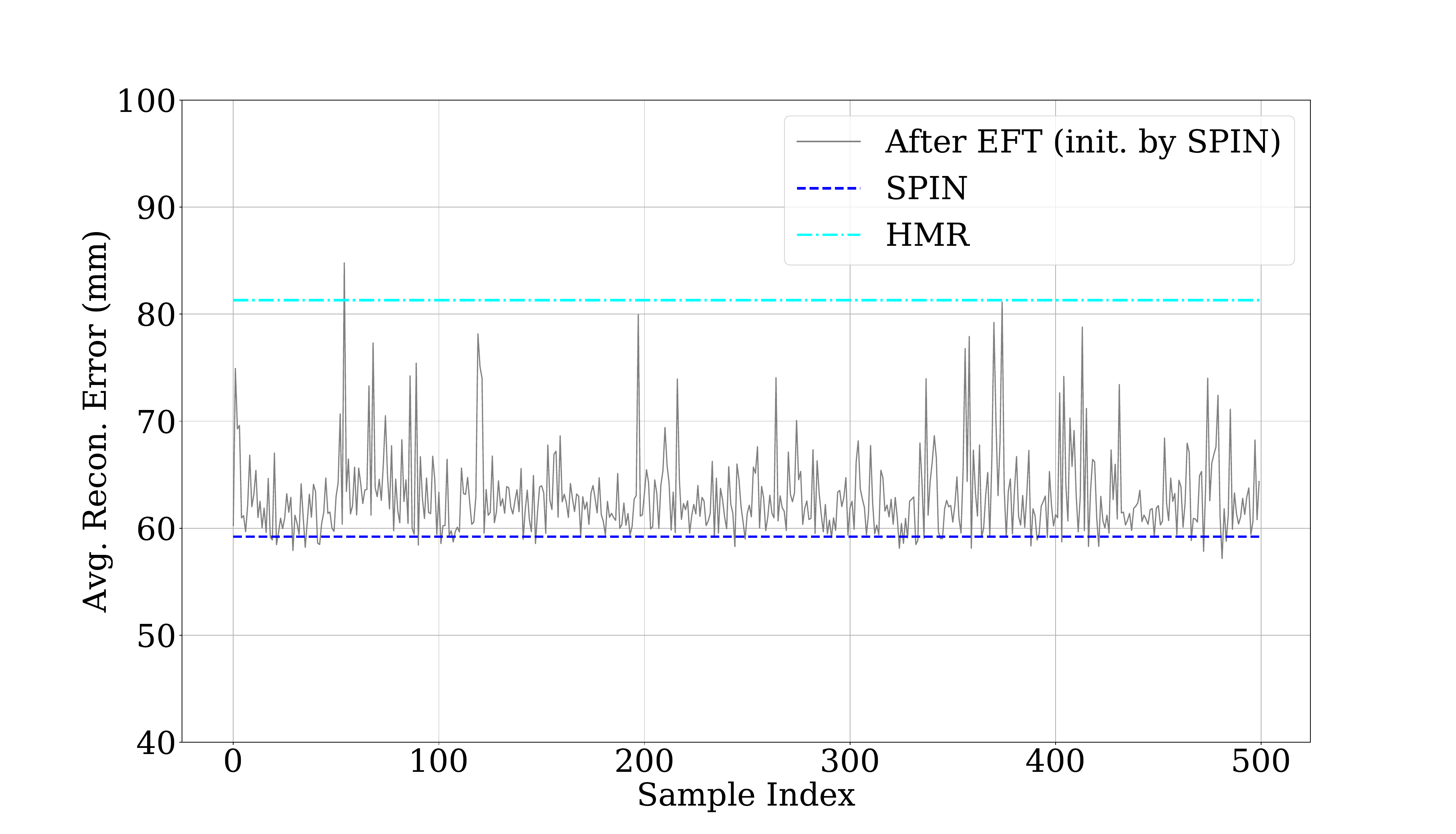}%
		\caption{Test error (PA-MPJPE) on 3DPW with the overfitted regressors by EFT on 500 different samples, for 20 (left) and 100 (right) iterations.}\label{fig:Exemplar_analysis_1}
	\end{figure}
	\subsection{Overfitting Analysis of EFT} \label{s:resultoverffing}
	As we do not use any explicit pose prior terms in our EFT optimization ($\gamma=0$ in~\cref{eq:exemplar_loss}), by overfitting to a single sample EFT could `break' the implicit pose prior originally captured in the regressor, potentially causing the results drifting away from the true poses in the later EFT iterations.
	To test this, we can check the generalization capabilities of the overfitted pose regressor from EFT procedure on each single sample, by evaluating the accuracy of the overfitted regressor on the entirety of the 3DPW  test set.
	We repeat this test for 500 different samples by using 20 and 100 EFT iterations respectively and report the results in \cref{fig:Exemplar_analysis_1}.
	The performance of SPIN~\cite{kolotouros2019spin} (blue line) and HMR~\cite{kanazawa2018end} (cyan line) are also shown for comparison.
	As can be noted, the overfitted regressors still perform very well overall, suggesting that the network \emph{retains} its good properties despite EFT\@.
	In particular, the performance is at least as good as the HMR baseline~\cite{kanazawa2018end}, and occasionally the overall performance improves after overfitting a single sample.
	The effect is different for different samples --- via inspection, we found that samples that are more likely to `disrupt' the regressor contain significant occlusions or annotation errors. Note that the overfitted network is discarded after applying EFT on a single example  --- this analysis is only meant to illustrate the effect of fine-tuning.
	
	\subsection{Qualitative Evaluation on Internet Videos}
	We demonstrate the performance of our 3D pose regressor models trained using our EFT datasets on various challenging real-wold Internet videos, containing cropping, blur, fast motion, multiple people, and other challenging effects. Example results are shown in the supp.~videos.
	\section{Discussion}
	
	We provide, via EFT, large-scale and high-quality pseudo-GT 3D pose annotations that are sufficient to train state-of-the art regressors. We expect out `EFT datasets' to be of particular interest to the research community, stripping the training of 3D pose regressors from complicated preprocessing or balancing techniques. Our 3D annotations on the popular 2D datasets such as COCO can also provide more opportunities to relate 3D human pose estimation with other computer vision tasks such as object detection and human-object interaction.
	
	\bibliographystyle{plain}\bibliography{egbib}
	
	\newpage
	\section*{Supplementary Material}
	\renewcommand\thesection{\Alph{section}}
	\renewcommand{\thesubsection}{\alph{subsection}}
	\setcounter{section}{0}
	
	In this supplementary material, we provide additional experiments and other details. We describe implementations of the corresponding ``gold-standard'' settings of SMPLify~\cite{Bogo2016} and EFT in \cref{ss:gold-standard}, and then further compare these two algorithms using an identical set of hyper parameters in \cref{ss:apple-to-apple}. More extensive quantitative evaluation on public benchmarks is provided in \cref{ss:quan-full-table}. We also include other details that were not provided in main manuscript due to space constraints.
	
	\section{Implementation Details of \emph{Gold-standard} Setting (main paper sec. 6.5)} \label{ss:gold-standard}
	Below we provide more details on implementations of ``gold-standard settings'' for each algorithm benchmarked in Sec. 6.5 of the main manuscript. 
	
	\paragraph{SMPLify.}
	For SMPLify, we use the setting from SPIN~\cite{kolotouros2019spin}\footnote{See the public SMPLify code from SPIN for further details : \url{https://github.com/nkolot/SPIN/blob/master/smplify/smplify.py}} with minor modifications. 
	For the data term, the 2D reprojection error is computed in pixel space of 224$\times$224 input images, with Geman-McClure robust function ($\sigma$=100). Prior terms include pose prior with GMM, angle prior to penalize unnatural bending, and shape prior, with respective weights of 4.78, 15.2, and 5.0.
	We use Adam optimizer with learning rate of $1\times10^{-2}$. The SMPL parameters $\bTheta = (\btheta, \bbeta)$ are first initialized with predictions of a pre-trained 3D pose regressor, same as in EFT. Then a multi-stage approach is employed by optimizing camera translation and body orientation first, and then all parts later.
	
	We made a few modifications to the original code: (1) we use weak perspective projection; (2) we do not use openpose input (see \cref{section:simplify_issue} for justification); and (3) we stop once the average 2d reprojection error is less than 3 pixels (max 50 iterations), which is the same setting as in EFT.
	
	\paragraph{EFT.}
	EFT uses the standard L2 loss for 2d keypoints data term (no robust loss function). Note that, as EFT follows the standard neural network training procedure, the reprojection loss is computed in a normalized image space (-1 to 1). 
	As mentioned in the main manuscript, the hip locations are ignored when computing reprojection errors, and we add a loss term to match the orientation of lower legs (with weight of $0.005$), encouraging the reconstruction of the orientation of the vector connecting the knee to the ankle to be similar to the one from 2D annotations. The key motivation underpinning the use of leg orientation loss is based on the observation that ankle annotations are particularly noisy, causing erroneous foreshortening during fitting process (see \cref{sec:noisy_gt}).
	
	We use Adam optimizer with learning rate $5\times10^{-6}$ (instead of 
	$5\times10^{-5}$ used for training the 3D pose regressor (also used in SPIN)), see more discussion below. We stop iterating once the average 2d error $\le$ 3 pixels (max 50 iterations), which corresponds the same setting as SMPLify.
	
	Note that, EFT computes the 2D reprojection error in a normalized space instead of the image pixel space as in the SMPLify. However, this does not make difference because Adam is re-scaling invariant. For example, we can compute the 2D reprojection error in either space after re-scaling 2D distance and adjusting the weights for the other terms accordingly, for which Adam provides the same updates. 
	
	
	\begin{table*}[t]
		\centering
		\scriptsize
		\begin{tabular}{l>{\columncolor{lightgray!70}}c|cccccc|ccccc}
			\toprule
			\multirow{2}{*}{\textbf{Models}}	&  & \multicolumn{6}{c|}{\textbf{SMPLify}} & \multicolumn{5}{c}{\textbf{EFT}} \\
			& \multirow{-2}{*}{\textbf{No post}} & $\gamma=0$ & $0.005$ & $0.01$ & $0.0427$ &  $0.1$ & $100.0^{\star}$ &
			\textbf{$\gamma=0$}  & $0.01$ & $0.1$ & $1.0$ & $100.0^{\star}$ \\
			\midrule
			H36M  & 146.20 & \textcolor{myred}{154.03} & \textcolor{myblue}{140.38} & \textcolor{myblue}{130.54}   & \textcolor{myblue}{123.19}  &  \textcolor{myblue}{125.55} & \textcolor{myblue}{119.67}
			& \textbf{\textcolor{myblue}{108.8}} & \textcolor{myblue}{108.92} & \textcolor{myblue}{109.84} & \textcolor{myblue}{114.89} & \textcolor{myblue}{137.79} \\
			$[\text{LSP}]_{\text{EFT}}$  & 84.90 & \textcolor{myblue}{76.77} & \textcolor{myblue}{70.91} & \textcolor{myblue}{70.09}   & \textcolor{myblue}{80.41}  &  \textcolor{myblue}{83.33} & \textcolor{myred}{87.72}
			& \textcolor{myblue}{68.00} & \textcolor{myblue}{67.85} & \textbf{\textcolor{myblue}{67.40}} & \textcolor{myblue}{68.43} & \textcolor{myblue}{82.01}\\
			$[\text{MPII}]_{\text{EFT}}$  & 69.10 & \textcolor{myblue}{66.16} & \textcolor{myblue}{64.25} & \textcolor{myblue}{64.53}   & \textcolor{myred}{73.42}  &  \textcolor{myred}{76.00} & \textcolor{myred}{85.59}
			& \textcolor{myblue}{63.08} & \textcolor{myblue}{63.06} & \textbf{\textcolor{myblue}{62.7}} & \textcolor{myblue}{63.01} & \textcolor{myblue}{68.61} \\
			
			CO-H36-M-3P  & 51.60 & \textcolor{myred}{51.72} & \textcolor{myblue}{50.90} & \textcolor{myred}{51.65}   & \textcolor{myred}{56.40}  &  \textcolor{myred}{57.64} & \textcolor{myred}{79.66}
			& \textcolor{myblue}{50.34} & \textcolor{myblue}{50.28}  & \textbf{\textcolor{myblue}{50.12}} & \textcolor{myblue}{50.27} & \textcolor{myblue}{51.50}\\
			\bottomrule
			
		\end{tabular}
		\caption{\textbf{Quantitative evaluation on 3DPW by using SMPLify and EFT as post-processing}, in PA-MPJPE errors ($mm$). Each row is the result with a pretrained model trained by different dataset (shown in the first column). We use the same data term for both methods. For SMPLify we test by varying $\gamma$ for the pose prior term (the default weight of SPIN is 0.0427). Similarly, we also test EFT by varying $\gamma$ for the neural network weight regularizer, where our original EFT uses $\gamma=0$. $\gamma={100.0}^\star$ means that we use $\gamma=100.0$ with a fixed 100 iterations without early stop. The bold numbers represent the cases with the best accuracy among tests. The numbers in blue indicate cases in which post-processing improves over the original predictions (which is always the case for the EFT post-processing), and the red color indicates loss in performance. ``CO+H36M+M+3P'' stands for ``$[\text{COCO-Train}]_{\text{EFT}}$ + H36M + $[\text{MI}]_{\text{EFT}}$ + 3DPW-T''
			\label{table:quant_apple2apple}}
	\end{table*}
	
	\paragraph{Justification of Different Learning Rates.} The learning rate of EFT $5\times10^{-6}$ should not be directly compared to the one in SMPLify $1\times10^{-2}$. In SMPLify, the optimizer directly updates the 85-dimensional vector of SMPL parameters, while in EFT it optimizes the \emph{weights} of the neural network (27 millions) that produces SMPL parameters via a series of non-linear computations (including Batch Normalization). In fact, no same learning rate can be used for both optimizations --- we confirmed that with learning rate of $1\times10^{-2}$ EFT changes SMPL parameters too much in each iteration, while with learning of $5\times10^{-6}$ SMPLify almost does not change SMPL parameters. For balancing, we instead observed reprojection error changes in the first few iterations, and found that the current learning rate settings produce similar step size in updating \emph{SMPL parameters}, which is also qualitatively confirmed by 3D visualizions.
	
	\paragraph{Discussion.}
	The performance of SMPLify is largely affected by hyper parameters such as weights between data term and prior terms, and also by the initialization states that motivates the use of a multi-stage technique. Given the excellent achievement of SPIN in the use of SMPLify on the same application with ours, we treat this setting as a ``gold-starndard''. The results shown in sec. 6.5 of our main manuscript should be understood as a comparison with their own best settings between EFT and SMPLify. In the following section, we perform an additional comparison in the same setting.

	\section{Comparison SMPLify and EFT in the Same Data Term Setting (main paper sec. 6.5)}\label{ss:apple-to-apple}
	
	In this section, we perform further comparison between EFT and SMPLify with the same set of hyper parameters. Concretely, the exact same data term is used for both methods, and an ablation study is performed by adding prior terms for each method with varying weights.
	
	\paragraph{Implementation Details.} 
	To employ the same data term for EFT and SMPLify, we make the following modifications. We remove Geman-McClure robust function in SMPLify and use standard L2 loss, computing the reprojection error in the normalized space (-1 to 1) as in EFT. For EFT, we remove the leg orientation data term. The exact same body keypoints, ignoring hips, are used in computing 2D reprojection loss for both methods. 
	We keep the same learning rates as in previous experiments, $1\times10^{-2}$ for SMPLify and EFT $5\times10^{-6}$.
	
	After performing a comparison with data-term only ($\gamma=0$ in \cref{table:quant_apple2apple}), we further investigate by adding prior terms of each method ($\gamma>0$). For SMPLify, we only consider the pose prior term, omitting the angle prior term and the shape prior term. The original weight for the pose prior term (4.78) of SMPLify is also adjusted to compensate the scaling change of the data term (i.e., $\gamma=0.0427$ in \cref{table:quant_apple2apple}). With varying $\gamma$, we analyze the performance changes of SMPLify.
	
	For EFT, we add the neural network weight regularizer term (the second term in eq. (3) of our main manuscript), by varying its weight $\gamma$.  
	
	
	
	\paragraph{Results.}
	As in sec. 6.5 and table 5 of our main manuscript, we carry out a quantitative evaluation by using the 3DPW dataset. As in the previous experiment, we use various regressors $\Phi$ to provide different initialization states, then use EFT and SMPLify for post-processing. 
	The results are shown in \cref{table:quant_apple2apple}, showing analogous conclusions to the result of our main paper (sec. 6.5 and table 5).

	When no prior term is used (i.e., $\gamma=0$), SMPLify produces mixed results, depending on the quality of initial states. For example, given a relatively accurate initialization (the last row), the post-processing makes the result slightly worse, and with intermediate models (the second and the third) it improves the final accuracy. With very poor initializations (the first low), the result becomes worse. 
	Adding pose prior on SMPLify also shows mixed performances. It is advantageous to have the pose prior when initializations are poor.  However, the use of prior degrades the accuracy if the initialization from regressor is accurate enough. 
	Note that, the post-processing performance changes quite significantly with small amount of weight changes (e.g., 51.72$mm$ to 57.64$mm$ in the last row, and more changes in other models). This particular result indicates that the weight $\gamma$ needs to be adjusted carefully to obtain the best performance in each scenario. 
	To explore further, we investigate the performance with an extremely strong $\gamma=100$ with fixed 100 iterations without early stopping (shown in $100.0^\star$), and, as expected, the results converge to a certain accuracy (around 86$mm$), presumably the mean pose defined by the pose prior term.

	Now, we see the results from EFT. When we use the data-term only $\gamma=0$, the same setting we use throughout our paper, it improves accuracy in \emph{all} cases. Furthermore, by performing experiments with varying $\gamma$, we confirmed that the results of EFT are not sensitive to the change of $\gamma$. Notably, adding prior does not degenerate the initial accuracy from regressor in all cases. This is still true even though an extremely large $\gamma=100$ is used with fixed 100 iterations (similar setting with SMPLify as shown in $100.0^\star$), where the regularizer term almost overshadows the data term. In this case, as expected, the results fall back to the ‘best guess’ of the pre-trained regressor network. The use of prior term sometimes helps to reduce the accuracy further, but the advantage is marginal. It shows less than 1$mm$ difference in most cases, providing a justification for our default setting with no pose prior term.
	
	\paragraph{Discussion}
	As demonstrated above, traditional fitting methods such as SMPLify suffer from the balancing issue between the data term and the ``view-agnostic'' 3D pose prior term. In contrast, EFT does not suffer from such issues, and we demonstrate that it can be applicable even without any prior terms. EFT effectively leverages better pose regressors, showing better performance in all cases.
	
	Note that this experiment is intended to explore the major difference between EFT and SMPLify, rather than finding the best settings for each method. The best SMPLify setting on 3DPW is not necessarily the best in other cases, and, often, it is difficult to determine the optimal set of hyper parameters. For example, we found that the use of prior terms or multi-stage optimization sometimes degrade the accuracy, but they can be still advantageous on other cases with poor initialization, such as OCHuman~\cite{pose2seg2019}. In general, the gold standard setting proposed in SMPLify~\cite{Bogo2016} and SPIN~\cite{kolotouros2019spin} is recommended for practitioners. Importantly, as demonstrated, EFT can provide a better option, showing better performance without complicated parameter tuning.
	
	\section{Further Evaluation on Standard Benchmarks (main paper sec. 6.2) }\label{ss:quan-full-table}
	
	The full evaluation results on the standard benchmarks including MPI-INF-3DHP~\cite{mehta2017monocular} are shown in \cref{table:quant_public_db_full}. All results are measured in terms of reconstruction errors (PA-MPJPE) in $mm$ after rigid alignment, following~\cite{kanazawa2018end, kolotouros2019spin}. We also report Per-Vertex Error in \cref{table:pve_3dpw}. Here, we highlight several noticeable results. 
	
	\paragraph{Evaluation on MPI-INF-3DHP~\cite{mehta2017monocular}.} The model trained with our EFT dataset is also competitive on the MPI-INF-3DHP dataset. The models trained by \emph{only}, $[\text{COCO-Part}]_{\text{EFT}}$ $[\text{COCO-Train}]_{\text{EFT}}$ outperform HMR~\cite{kanazawa2018end}. Combining 3D datasets with our EFT dataset improves the performance further.

	\paragraph{H36M Protocol-1.} We also report quantitative results on H36m protocol 1. In this evaluation, we use all available camera views in H36M testing set. The overall errors are slightly higher than protocol-2 (frontal camera only), but the general tendency is similar to the results of H36M protocol-2.

	\paragraph{A Baseline Model Trained with 2D Loss Only.} As a na\"ive baseline method, we train a 3D pose regressor by just using 2D annotations (that is, with only 2D keypoint loss) without including other 3D losses from real or pseudo-ground truth 3D datasets. The results are shown as $[\text{COCO-Part}]_{\text{2D}}$ and $[\text{COCO-Train}]_{\text{2D}}$ in \cref{table:quant_public_db_full}. As expected, the model cannot estimate accurate 3D pose, showing very poor performance. However, interestingly, we found the 2D projection of the estimated 3D keypoints to still be quite accurately aligned to the target individual's 2D joints.

	
	\begin{table}[t]
		\centering
		\footnotesize
		\begin{tabular}{l|c>{\columncolor{lightgray!70}}c}
			\toprule
			\textbf{Datasets}	&  \textbf{MPJPE	$\downarrow$} &  \textbf{PVE 	$\downarrow$}\\
			\midrule
			HMR  & 81.3 & 142.7 \\
			SPIN  & 59.2 & 129.2  \\
			VIBE & 56.5 & 99.1 \\
			\midrule
			$[\text{COCO-Part}]_{\text{EFT}}$ & 59.0 & 115.3 \\
			$[\text{COCO-Train}]_{\text{EFT}}$ & 57.5 & 112.5 \\
			$[\text{COCO-Train}]_{\text{EFT}}$ + H36M & 55.5 & 107.6 \\
			$[\text{COCO-Train}]_{\text{EFT}}$ + H36M + MI & 54.7 & 106.1 \\
			$[\text{COCO-Train}]_{\text{EFT}}$ + H36M + MI + 3DPW & 51.6 & 97.5 \\
			\bottomrule
		\end{tabular}
		\caption{Per-Vertex-Error on 3DPW dataset.
			\vspace{-5pt}
			\label{table:pve_3dpw}}
	\end{table}
	
	\paragraph{PVE Error:} We report Per-Vertex Error in Table~\ref{table:pve_3dpw} between the ground truth mesh vertices and the vertices from predictions. While PVE error would be affected by shape parameters which is not the major focus of our work, its tendency is aligned to the performance quantified by MPJPE errors, showing lower PVE errors with the model with lower MPJPE errors.

	\section{Training Models with Auxiliary Inputs (main paper sec. 6.3) }
	In addition to the results shown in section 6.3, we explore the performance of the models trained with other auxiliary inputs. We analyze the results of the models trained with two different segmentation maps: Panoptic Segmentation~\cite{kirillov2019panoptic} that has both things and stuff classes and PointRend~\cite{kirillov2020pointrend} that has only thing classes but with more accurate contours. The results are shown in \cref{table:quant-auxiliary-input}, along with the original results by RGB only and the results with DensePose~\cite{guler2018densepose}. As can be seen, the color-coded segmentation encodings do not noticeably improve performance, compared to the results with DensePose.

	
	\begin{table*}[t]
		\centering
		\rowcolors{1}{}{lightgray}
		\begin{tabular}{lccccc}
			\toprule
			\textbf{Dataset Combinations} & $[\text{X}]_{\text{EFT}}$ & \textbf{H36M} & \textbf{MPI-INF} & \textbf{PanopticDB}  & \textbf{3DPW}  \\
			\midrule
			$[\text{X}]_{\text{EFT}}$    & 100\% &  0 & 0 & 0  & 0\\
			$[\text{X}]_{\text{EFT}}$ + H36m & 60\% &  40\% & 0 & 0  & 0\\
			$[\text{X}]_{\text{EFT}}$ + H36m + MPI-INF   & 30\% &  50\% & 20\% & 0  & 0 \\
			$[\text{X}]_{\text{EFT}}$ + H36m + MPI-INF + 3DPW & 30\% &  40\% & 10\% & 0  & 20\% \\
			\bottomrule
		\end{tabular}
		\caption{
			Training data sampling ratios across datasets. 
			\label{table:dataset-ratio}}
	\end{table*}
	
	\begin{table}[t]
		\centering
		\begin{tabular}{l|c}
			\toprule
			\textbf{Dataset Combinations} & \textbf{3DPW}  \\
			\midrule
			$[\text{COCO-Part}]_{\text{EFT}}$ &  59.0\\
			$[\text{COCO-Part}]_{\text{EFT}}$ w/ DensePose   & \textbf{57.2}  \\
			$[\text{COCO-Part}]_{\text{EFT}}$ w/ PanopticSeg    & 59.1  \\
			$[\text{COCO-Part}]_{\text{EFT}}$ w/ PointRend  & 58.9  \\
			\midrule
			$[\text{COCO-Train}]_{\text{EFT}}$ &  57.5 \\
			$[\text{COCO-Train}]_{\text{EFT}}$ w/ DensePose  & \textbf{56.1} \\
			$[\text{COCO-Train}]_{\text{EFT}}$ w/ PanopticSeg  & 57.4  \\
			$[\text{COCO-Train}]_{\text{EFT}}$ w/ PointRend  & 57.6  \\
			\bottomrule
		\end{tabular}
		\caption{
			Quantitative evaluation on 3DPW (PA-MPJPE in $mm$) by learning models with auxiliary inputs: DensePose~\cite{guler2018densepose}, segmentation map by Panoptic Segmentation~\cite{kirillov2019panoptic}, and segmentation map by PointRend~\cite{kirillov2020pointrend}
			\label{table:quant-auxiliary-input}}
	\end{table}
	
	\section{Overfitting Analysis of EFT: Examples (main paper sec. 6.6)}
\begin{figure}[t]
		\centering
		\includegraphics[trim=0 0 0 0,clip,width=\linewidth]{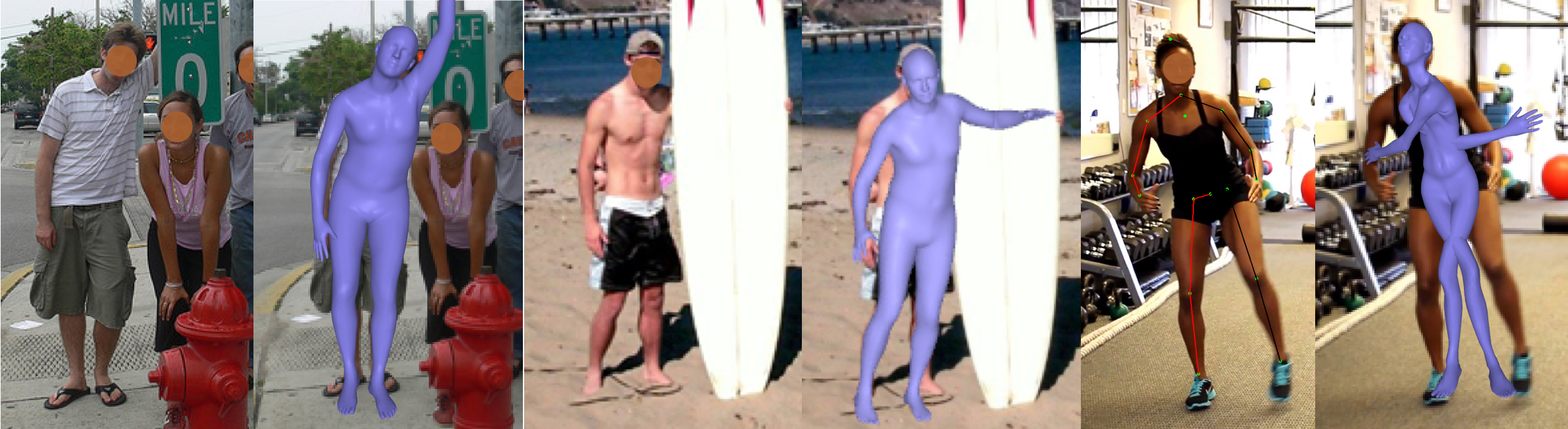}
		\caption{%
			Example samples that cause significant changes in the 3DPW testing error in our EFT overfitting test (main paper sec. 6.6)\@.
			The left two examples have annotations on occluded body parts, and the rightest example has incorrect annotations (left-right swap).}\label{fig:exemplar_analysis_vis}
	\end{figure}

	As shown in \cref{fig:Exemplar_analysis_1} of our main manuscript, overfitting the network to individual samples with EFT usually has a small effect on the \emph{overall} regression performance, suggesting that the network retains its good properties despite fine-tuning on exemplars. 
	The effect is different for different samples, and we show the examples with a strong effect in \cref{fig:exemplar_analysis_vis} that contain significant occlusions or annotation errors.

	\section{More Qualitative Comparisons Between EFT vs. SMPLIfy (main paper sec. 4)}
\begin{figure*}[t]
		\centering
		\includegraphics[trim=0 0 0 0,clip,width=\linewidth]{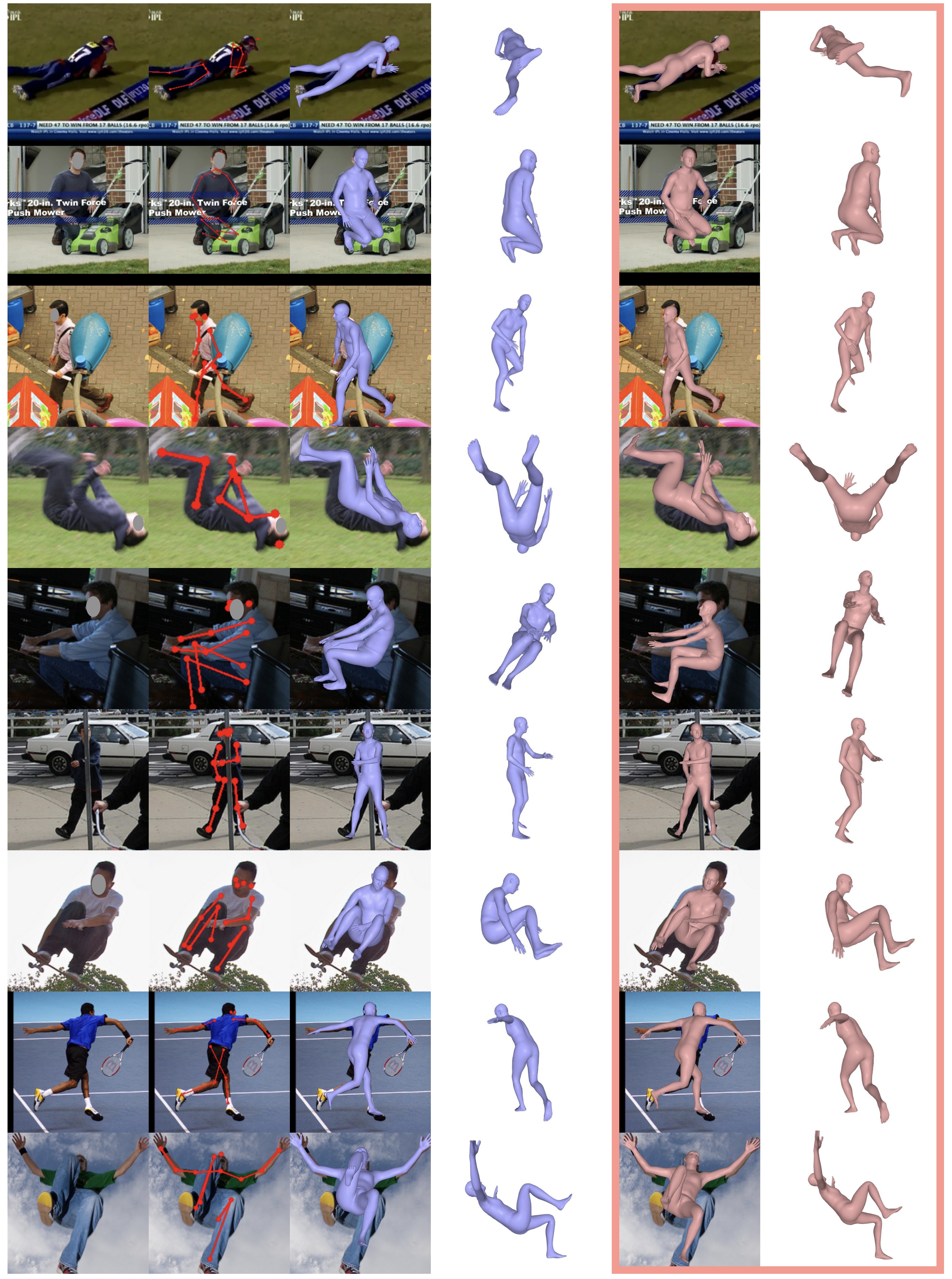}
		\caption{%
			The samples where SMPLify outputs are favored by all three annotators. The blue meshes are the results by EFT and the pink meshes are results by SMPLify. }\label{fig:amt_vote0_1}
	\end{figure*}

\begin{figure*}[t]
		\centering
		\includegraphics[trim=0 0 0 0,clip,width=\linewidth]{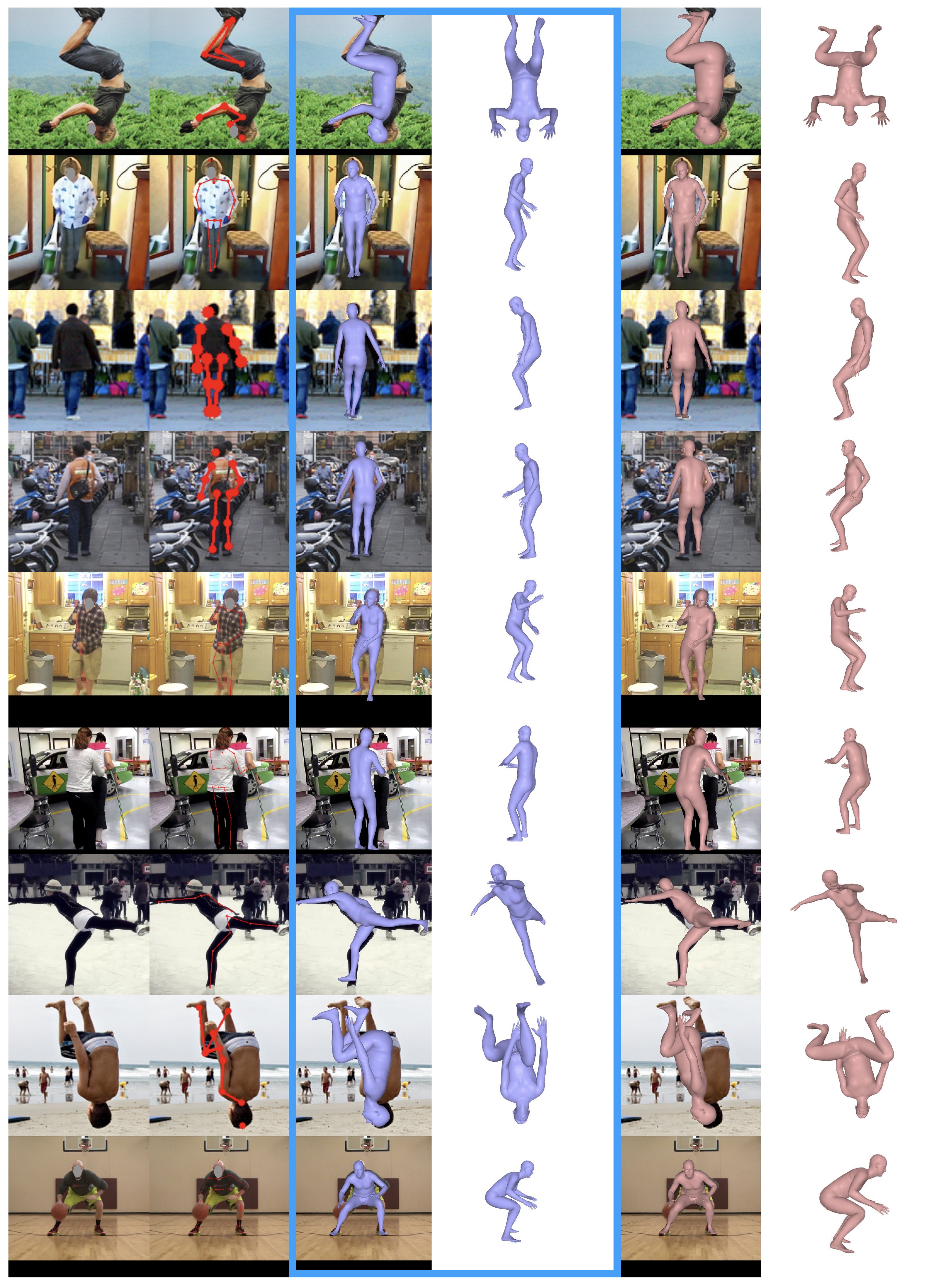}
		\caption{%
			The samples where our EFT outputs are favored by all three annotators. The blue meshes are the results by EFT and the pink meshes are results by SMPLify.}\label{fig:amt_vote3_1}
	\end{figure*}

	As addressed in our main paper, our EFT method was preferred \textbf{61.8\%} of the times with a majority of at least 2 votes out of 3 in Amazon Mechanical Turk (AMT) study. Here, we visualize the examples with 0 vote and 3 votes, as shown in \cref{fig:amt_vote0_1} and \ref{fig:amt_vote3_1}. Note that in our AMT study, we show the meshes over white backgrounds to reduce the bias cased by 2D localization quality. 
	
	\paragraph{When annotators prefer SMPLify?}
	There are 47 / 500 samples where SMPLify outputs are favored by all three annotators, and examples are shown in \cref{fig:amt_vote0_1}. Surprisingly the difference is quite minimal with no obvious failure patterns. 
	
	\paragraph{When annotators prefer EFT?}
	There were 132 / 500 samples where our EFT outputs were favored by all three annotators. Examples are shown in \cref{fig:amt_vote3_1}. In most cases, EFT produces more convincing 3D poses  leveraging the learned pose prior conditioned on the target raw image, while SMPLify tends to produce outputs similar to its prior poses (e.g., knees tend to be bent).

	\section{Computation Time}
	We compare computation time between our EFT and SMPLify processes. The computation times are computed for their \emph{gold-standard settings} by using a single GeForce RTX 2080 GPU. 
	
	\paragraph{EFT.}
	A single EFT iteration takes about 0.04 sec, and a whole EFT process for a sample data (including reloading the pre-trained network model) takes about \textbf{0.82 sec} with 20 iterations.
	
	\paragraph{SMPLify.}
	Camera optimization takes about 0.01 sec per iteration, and, thus, it takes 0.5 sec per sample (with 50 iterations). Optimizing body pose takes about 0.02 sec/iteration, and it takes 1 sec with 50 iterations. Thus whole SMPLify process takes about \textbf{1.5 sec} per sample with 50 iterations.

	\section{Noisy 2D keypoint issue in GT data (Implementation  details in sec. 3)}
	\label{sec:noisy_gt}
\begin{figure}[t]
		\includegraphics[width=\textwidth]{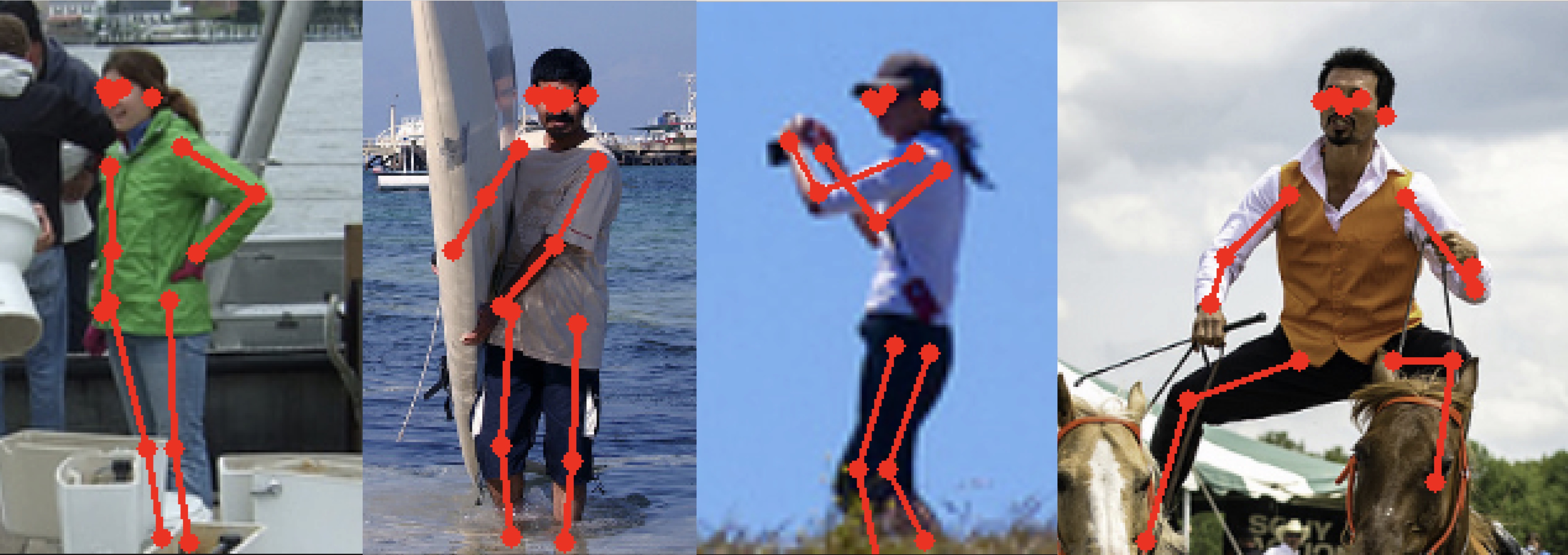}
		\caption{Noisy localization issue of 2D GTs, particularly on hips and ankle joints.
			\label{fig:noisy_2dkeypoint}}
	\end{figure}

	The keypoints of hips and ankles are more difficult to correctly localize by annotators than other keypoints due to loose clothing and occlusions. This can be also observed by checking the standard deviation of annotated 2D keypoints in COCO dataset (called OKS sigma values)~\footnote{\url{https://github.com/cocodataset/cocoapi/blob/8c9bcc3cf640524c4c20a9c40e89cb6a2f2fa0e9/PythonAPI/pycocotools/cocoeval.py}}, where hips (1.07) and ankles (0.89) have higher values than shoulders (0.79) and wrists (0.62). We show examples from COCO dataset in \cref{fig:noisy_2dkeypoint}, where in the annotations the lower leg lengths are shorter than it should be. We empirically found that such noisy 2D localization causes artifacts in 3D, motivating us to use 2d orientations for the lower leg parts rather than GT locations.

	\section{Statistics of our EFT datasets}
\begin{figure}[t]
		\vspace{-20pt} 
		\centering
		\includegraphics[width = 0.48\linewidth]{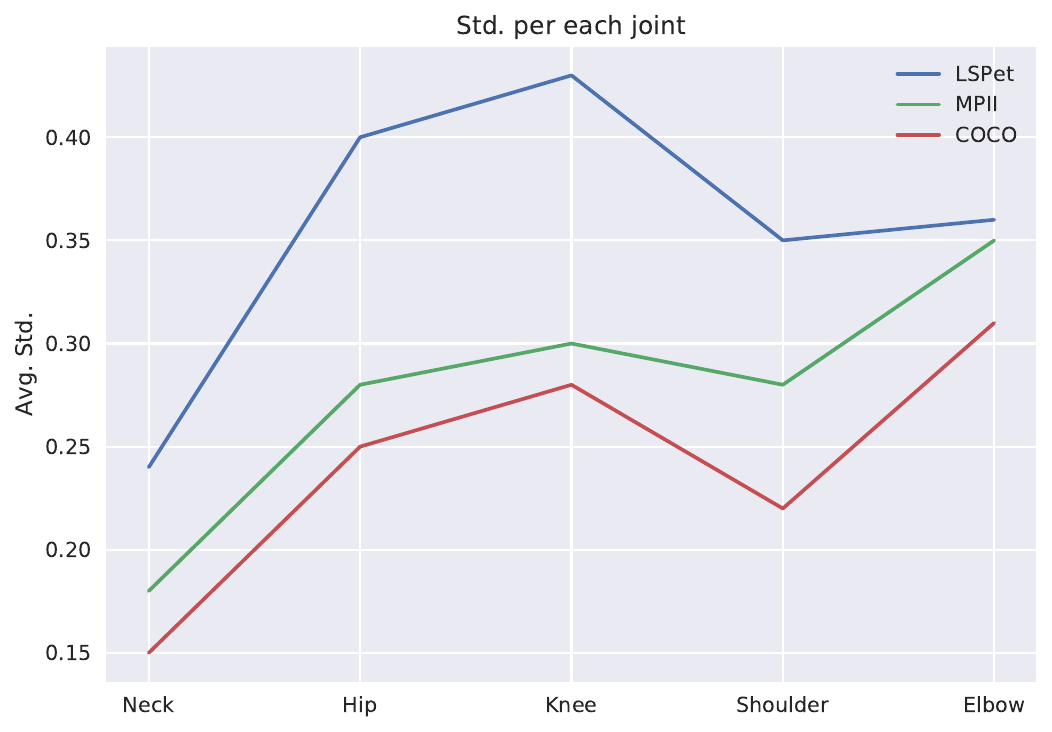}
		\includegraphics[width = 0.48\linewidth]{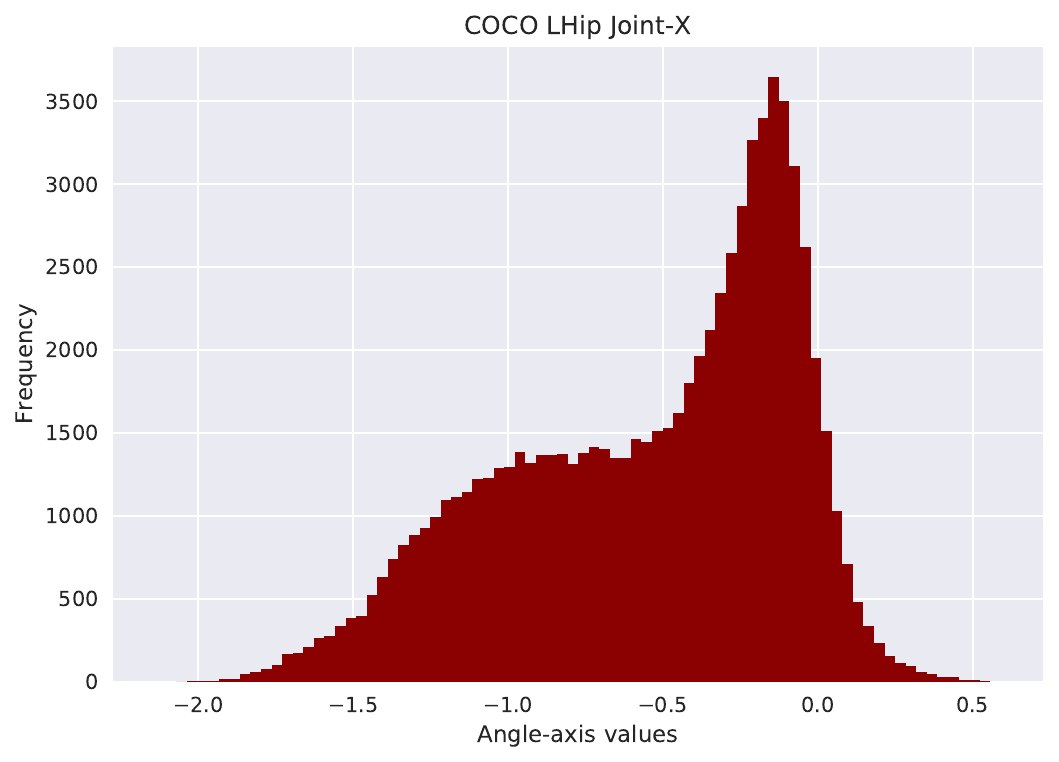}\\
		\includegraphics[width = 0.48\linewidth]{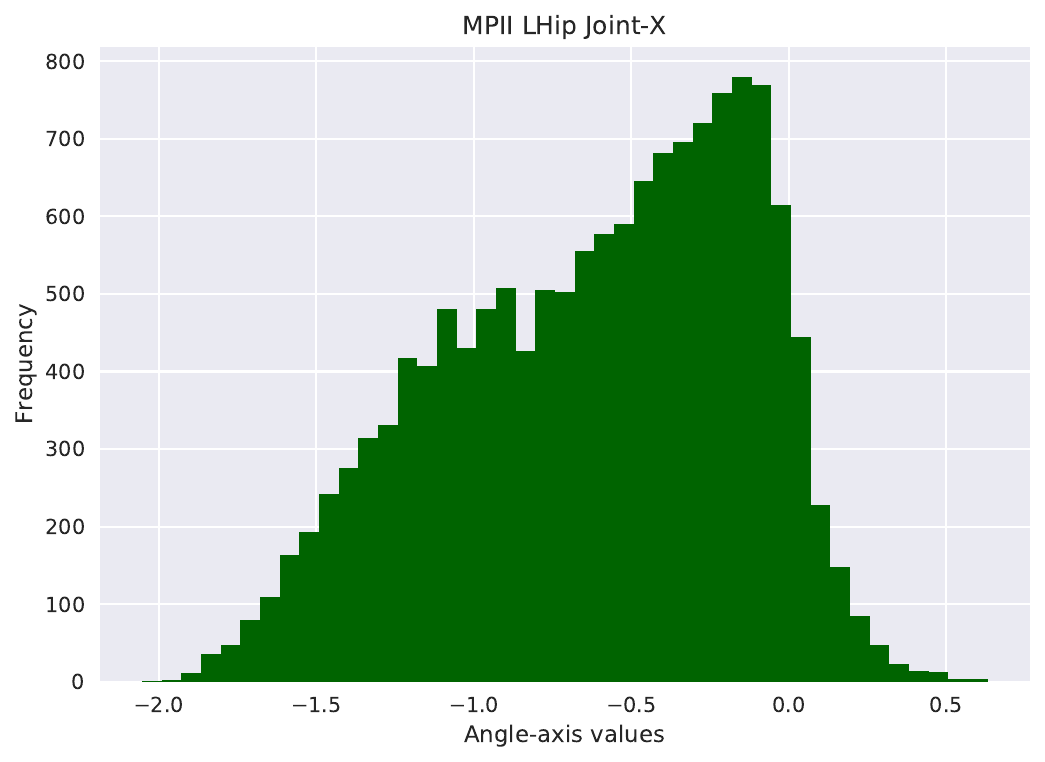}
		\includegraphics[width = 0.48\linewidth]{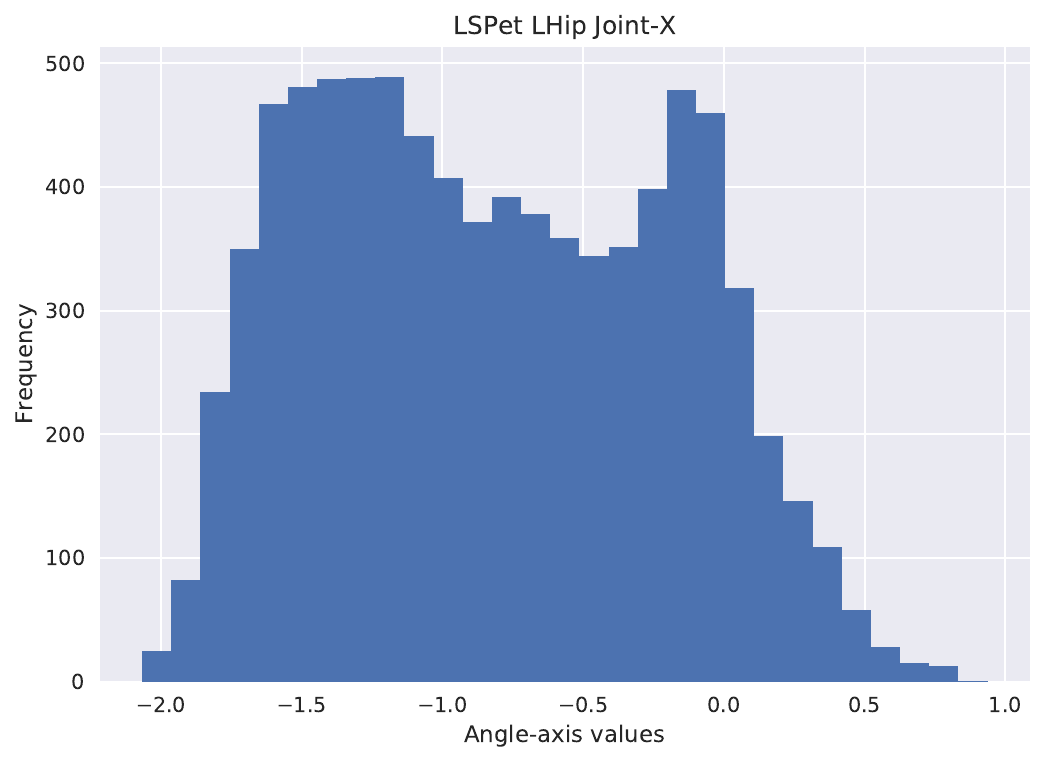}
		\caption{We compare the 3D pose parameter variations among COCO (red), MPII (green), and LSPet (blue). (Top-left) standard deviations of X-axis in angle-axis representation per selected joints. (Others) the histograms of the X-axis angular values of LHip joints.
			\label{fig:stats_dbs}
		}
	\end{figure}

	Since we now have high quality pseudo 3D annotations, we can check and compare the 3D pose variations among datasets. In \cref{fig:stats_dbs}, we compare the variations of 3D pose parameters of the pseudo GT data produced from three in-the-wild datasets, COCO (red), MPII (green), and LSPet (blue). The top-left figure shows the standard deviation of pose parameters (we choose the X-axis in angle-axis representation) for each body joint, including necks, hips, shoulders, and elbows. We can see that the LSPet dataset has the most diverse pose variations compared to MPII and COCO. In the later figures, we visualize the histograms of the pose parameters of a particular joint (Left Hip), to see the pose variations and distributions of all samples. As shown, the 3D poses of LSPet have higher portion of samples that have large angle axis values (less than -1.0), showing that it has more gymnastic poses. In contrast, the majority of COCO dataset has the values around zeros, we is closer to the rest standing pose.
	
	\section{Sampling Ratios across Datasets During Training}
	
	Following the approach of SPIN~\cite{kolotouros2019spin}, we use fixed sampling ratio across datasets while training the 3D pose regressor. The ratios used in our experiments are shown in \cref{table:dataset-ratio}. $[\text{X}]_{\text{EFT}}$ denotes our EFT dataset. If multiple EFT datasets are used (e.g., COCO and MPII), we sample from them uniformly.

	\begin{figure}[t]
		\centering
		\includegraphics[trim=0 0 0 0,clip,width=\linewidth]{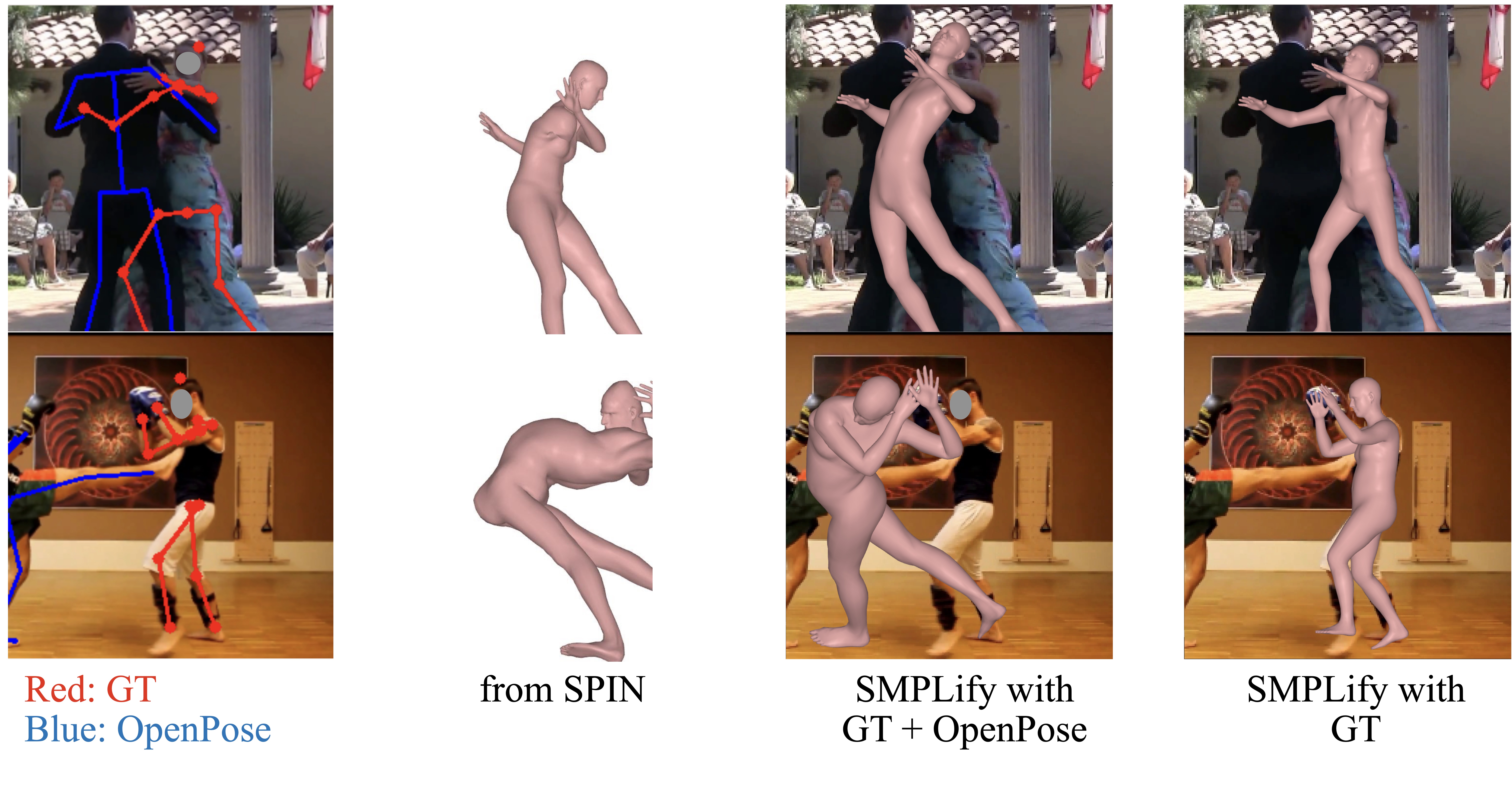}
		\caption{In SMPLify implementation of SPIN~\cite{kolotouros2019spin}, sometimes the OpenPose estimations (blue 2D keypoints) from different individuals are incorrectly associated to the target person (with red GT 2d keypoints). Thus, we completely ignore OpenPose estimation during our SMPLify process. (1st column) input images with GT annotaiton (red) and OpenPose estimation (blue); (2nd column) Fitting output produced from SPIN~\cite{kolotouros2019spin} (the publicly available data); (3rd column) our SMPLify output by using both GT and incorrectly associated OpenPose estimation; (4th column) our SMPLify output by using GT 2D keypoints without using OpenPose output. }\label{fig:spin_fit_issues}
	\end{figure}

	\section{Analysis on the Failures of SMPLify Fittings in SPIN}
	\label{section:simplify_issue}
	We found erroneous cases in the publicly available SMPL fitting outputs from SPIN~\cite{kolotouros2019spin} that are related to its SMPLify~\cite{Bogo2016} implementation. In SPIN, both 2D keypoint annotations and OpenPose~\cite{cao2018openpose} estimations are used for SMPLify, assuming that the OpenPose estimations have better localization accuracy if OpenPose is applied to the training data. However, we found many cases where the OpenPose outputs from different individuals are mistakenly associated to the target person, as shown in \cref{fig:spin_fit_issues}. Due to this reason, we exclude the OpenPose estimations and only use the GT annotations for our SMPLify process.

	\begin{figure}[t]
		\centering
		\includegraphics[trim=0 0 0 0,clip,width=\linewidth]{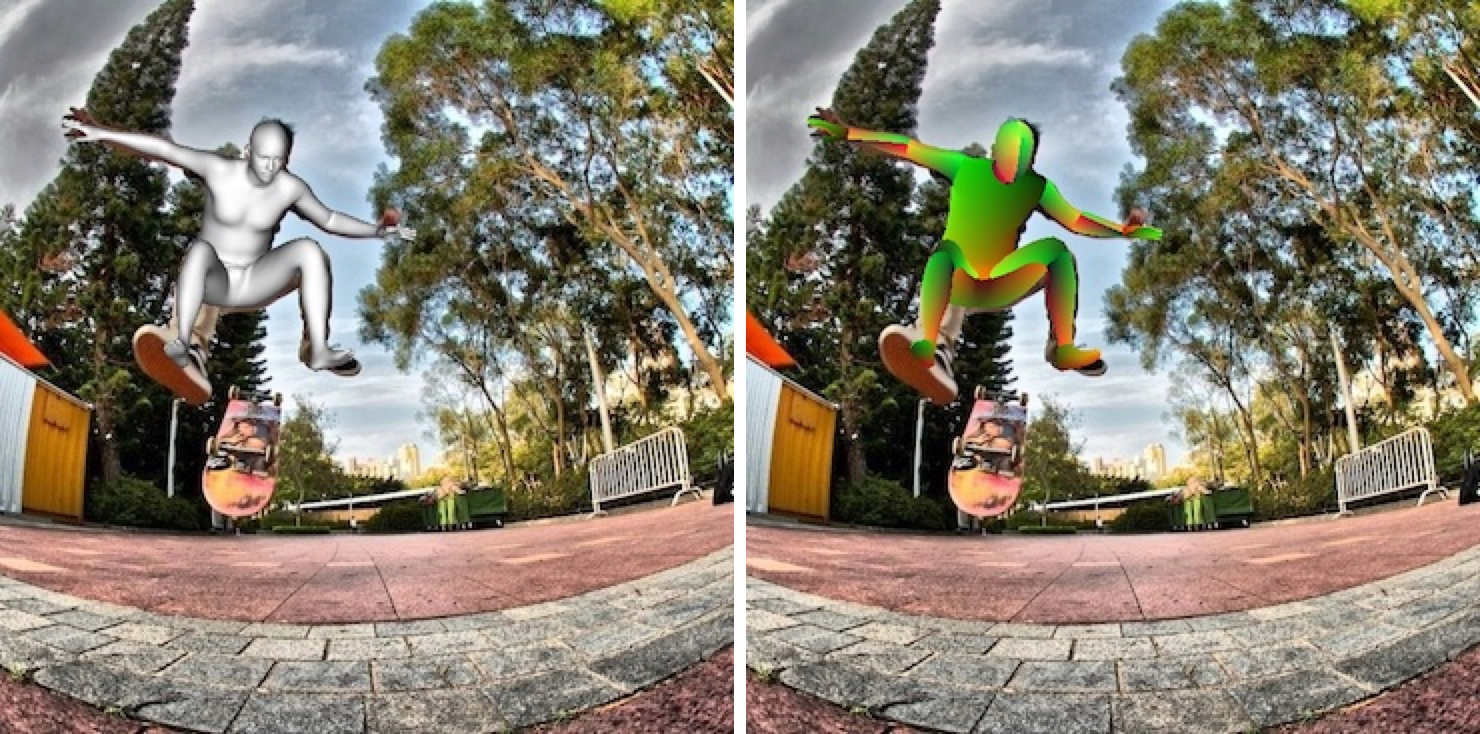}
		\caption{%
			Automatically generated densepose annotation}\label{fig:densepose_render}
	\end{figure}

	\begin{table*}[t]
		\centering
		\rowcolors{1}{}{lightgray}
		\begin{tabular}{lcccc}
			\toprule
			\textbf{Previous method} & \textbf{3DPW} $\downarrow$ & \textbf{H36M (P1)} $\downarrow$ & \textbf{H36M (P2)} $\downarrow$ & \textbf{MPI-INF-3DHP} $\downarrow$\\
			\midrule
			HMR~\cite{kanazawa2018end} & 81.3 & 58.1 & 56.8 & 89.8 \\
			DenseRaC~\cite{xu2019denserac}  & --- & --- & 48.0 & ---  \\
			DSD~\cite{sun2019human} & 75.0 & --- & 44.3 & ---\\
			HoloPose~\cite{guler2019holopose} & --- & --- & 50.6 & ---\\
			HoloPose (w/ post-proc.)~\cite{guler2019holopose} & --- & --- & 46.5 & ---\\
			SPIN~\cite{kolotouros2019spin} & 59.2 & --- & 41.1 & 67.5\\
			HMR (temporal)~\cite{humanMotionKanazawa19} & 72.6 & --- & 56.9 & ---\\
			Sim2Real (temporal)~\cite{doersch2019sim2real} &  74.7 & --- & --- & --- \\
			Arnab et al. (temporal)~\cite{arnab2019exploiting}    & 72.2 & --- &  54.3 &---\\
			DSD (temporal)~\cite{sun2019human} & 69.5 & --- & 42.4 & ---\\
			VIBE (temporal)~\cite{kocabas2019vibe} & 56.5 & 44.2 & 41.4 & 63.4\\
			\midrule
			\rowcolor{white}
			\multicolumn{3}{l}{\textbf{Straight 3D supervision and pseudo-ground truth from EFT}}\\
			\midrule
			H36M  & 146.2  & 56.1 & 53.3  & 147.4 \\
			MPI-INF-3DHP (MI) & 127.7   & 116.7 & 110.7  & 93.2 \\
			3DPW (Train)    & 91.0   & 139.8 & 131.4 & 125.2 \\
			\midrule
			$[\text{LSP}]_{\text{EFT}}$ & 84.9  & 91.6 & 87.2 & 97.1\\
			$[\text{MPII}]_{\text{EFT}}$    & 69.1  & 79.9 & 78.9 & 90.4\\
			$[\text{PoseTrack}]_{\text{EFT}}$  (PT)   & 72.6  & 91.1 & 88.6 & 98.7\\
			$[\text{COCO-Part}]_{\text{EFT}}$   & \textbf{59.0}  & 70.4 & 66.9 & 83.6 \\
			$[\text{COCO-Train}]_{\text{EFT}}$    & \textbf{57.5}& 67.2  & 64.2 & 80.2\\
			$[\text{COCO-Part}]_{\text{EFT}}$ + H36M & \textbf{57.8} & 47.8 & 45.0 & 77.5\\
			$[\text{COCO-Train}]_{\text{EFT}}$ + H36M & \textbf{55.5} & 48.7 & 46.2 & 76.1\\
			$[\text{COCO-Train}]_{\text{EFT}}$ + $[\text{PoseTrack}]_{\text{EFT}}$ & \textbf{56.5} & 67.1 & 63.5 & 79.8\\
			
			$[\text{COCO-Part}]_{\text{EFT}}$ + H36M + MI   & \textbf{56.9} & 47.7 & 45.4 & 67.6\\
			$[\text{COCO-Train}]_{\text{EFT}}$ + H36M + MI    & \textbf{54.7}  & 47.8 & 44.9 & 68.1\\
			
			$[\text{CO-Tr}]_{\text{EFT}}$ + $[\text{PT}]_{\text{EFT}}$ + $[\text{LSP}]_{\text{EFT}}$+ H36m + MI  & \textbf{54.3}  & 48.6 & 46.0  & 68.5\\
			
			$[\text{CO-Tr}]_{\text{EFT}}$ + $[\text{PT}]_{\text{EFT}}$ + $[\text{LSP}]_{\text{EFT}}$+ $[\text{OCH}]_{\text{EFT}}$ + H36m + MI  & \textbf{55.0}  & 47.5 & 44.8  & 67.4\\
			
			$[\text{COCO-Train}]_{\text{EFT}}$ + H36M + MI + 3DPW & \textbf{51.6}  & 47.1 & 44.0  & 67.5\\
			$[\text{CO-Tr}]_{\text{EFT}}$ + $[\text{PT}]_{\text{EFT}}$ + $[\text{LSP}]_{\text{EFT}}$+ H36m + MI +3DPW  & \textbf{52.4}  & 46.8 & 43.9  & 68.3\\
			
			\midrule
			\rowcolor{white}
			\multicolumn{3}{l}{\textbf{with Crop Augmentation}}\\
			\midrule
			$[\text{CoCO-Train + PT + LSP}]^{ca}_{\text{EFT}}$ + H36M + MI    & \textbf{53.6}  & 48.2 & 45.6 & 68.0\\
			$[\text{CoCO-Train + PT + LSP + OCH}]^{ca}_{\text{EFT}}$ + H36M + MI    & \textbf{54.6}  & 48.4 & 45.4 & 67.9 \\
			
			\midrule
			\rowcolor{white}
			\multicolumn{1}{l}{\textbf{Alternative Approaches}}\\
			\midrule
			$[\text{COCO-Part}]_{\text{SPIN~\cite{kolotouros2019spin}}}$ & 70.9  & 93.3 & 78.2 & 93.2\\
			$[\text{COCO-Part}]_{\text{SMPLify}}$   & 72.7 & 87.6 & 81.0 & 98.0\\
			
			$[\text{COCO-Part}]_{\text{2D}}$ & 228.1 &  236.4 & 213.7 & 218.8\\
			$[\text{COCO-Train}]_{\text{2D}}$ & 192.1 & 198.4 & 181.1 & 195.3\\
			$[\text{COCO-Part}]_{\text{2D}}$ + H36M + MI & 72.6 &  60.7 & 56.9 & 80.9\\

			\bottomrule
		\end{tabular}
		\caption{
			\textbf{Quantitative evaluation on the 3DPW, H36M protocol-1 (by using all views), H36M protocol-2 (by using frontal views), and MPI-INF-3DHP datasets.} Reconstruction errors are computed by PA-MPJPE and  reported in $mm$ after alignment by a rigid transformation. 
			\label{table:quant_public_db_full}}
	\end{table*}

	\section{Potential Applications for EFT Datasets}
	
	Our EFT datasets provide high-quality pseudo-GT 3D pose data on in-the-wild images. This can potentially open up several new research opportunities. 
	
	\paragraph{DensePose Annotations.}
	We can automatically render DensePose~\cite{guler2018densepose} annotations from our pseudo GT data. Examples are shown in ~\cref{fig:densepose_render}. As a potential research direction, these automatically generated densepose annotations can be used for training.
	
\begin{figure*}[t]
		\centering
		\includegraphics[trim=0 0 0 0,clip,width=\linewidth]{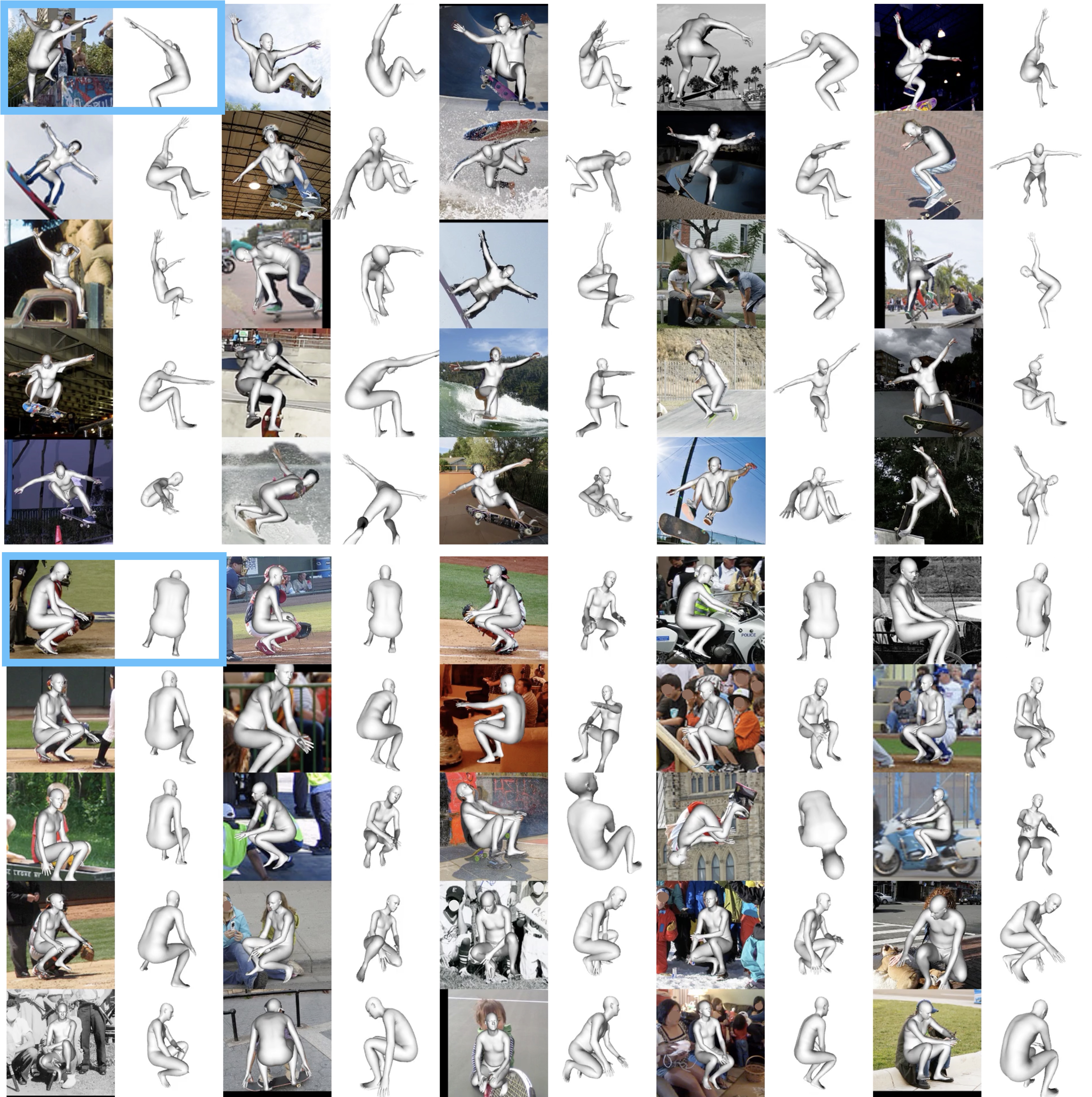}
		\caption{%
			Nearest Neighbor 3D Pose Search in COCO. We can find similar 3D poses by using the one in the top-left as a query (shown in blue boxes) in COCO EFT Dataset.}\label{fig:knn_coco}
	\end{figure*}

	\paragraph{Nearest Neighbor Search.}
	Our pseudo-GT datasets allow us to further analyse diverse 3D human poses in the context of outdoor environments. For example, by using a simple nearest neighbor we can find similar 3D human poses in the COCO dataset. The examples are shown in \cref{fig:knn_coco}. 

\end{document}